\documentclass[letterpaper]{article} 
\usepackage{aaai2027}  
\usepackage[hyphens]{url}  
\usepackage{graphicx} 
\usepackage{natbib}  
\usepackage{caption} 
\usepackage{algorithm}
\usepackage{algorithmic}

\usepackage{newfloat}
\usepackage{listings}
\DeclareCaptionStyle{ruled}{labelfont=normalfont,labelsep=colon,strut=off} 
\floatstyle{ruled}
\newfloat{listing}{tb}{lst}{}
\floatname{listing}{Listing}

\usepackage{booktabs}
\usepackage{amssymb}
\usepackage{multirow}
\usepackage{graphicx}
\usepackage{amsmath}

\nocopyright

\newcommand{\projname}{\textsc{Practice}}
\title{\projname{}: From Experience to Expertise in Self-Evolving Embodied Agents}

\author{
Ziyi Bai\textsuperscript{\rm 1}\equalcontrib,
Siqi Li\textsuperscript{\rm 2,3}\equalcontrib,
Tinglei Huang\textsuperscript{\rm 2},
Börje F. Karlsson\textsuperscript{\rm 1}\corresponding
}

\affiliations{
\textsuperscript{\rm 1}Beijing Academy of Artificial Intelligence (BAAI)\\
\textsuperscript{\rm 2}Institute of Software, Chinese Academy of Sciences\\
\textsuperscript{\rm 3}University of Chinese Academy of Sciences
}

\begin{document}

\maketitle

\begin{abstract}
Recent studies have shown that multimodal large language models (MLLMs) can serve as embodied agents, translating language instructions and visual observations into executable plans. 
However, building agents that can continually improve through interaction and rapidly adapt to their environments remains challenging. 
Summing up experience from past interaction trajectories provides a promising solution, but existing experience-based methods often rely on manually designed prompting workflows to extract and update skills. 
Such fixed procedures may struggle to learn updated skills from new and diverse experiences.
We introduce \projname{}, which trains a skill learner to discover and maintain a persistent skill library from past interaction trajectories while keeping the task executor frozen. 
Given the historical accumulated skills and incoming trajectories, the skill learner produces structured batch-edits that add, refine, merge, or remove skills, and then hierarchical consolidate all collected edits into a consistent updated skill library. 
We train the learner with a two-stage curriculum. First, it learns basic skill generation and library maintenance from oracle trajectories. 
Then, by contrasting successful and failed trajectories from heterogeneous executors on the same tasks, it learn to identify invalid action patterns and recovery strategies. 
Finally, we apply online skill-edit distillation to align the skill learner with a stronger teacher on its current edit distribution to further improves the policy. 
Experiments demonstrate that a compact skill learner delivers consistent performance improvements across successive library-update rounds for multiple frozen executors. 
On EB-ALFRED and EB-Habitat, \projname{} further outperforms the strongest experience-based baselines by 9.7 and 2.6 percentage points respectively. Project resources are publicly available at: 
\url{https://baai-agents.github.io/PRACTICE}.
\end{abstract}




\section{Introduction}



Embodied task planning has emerged as an important testbed for general-purpose agents, requiring them to interpret human instructions, decompose complex goals into executable steps, and accomplish those goals through interaction with the environment
~\cite{yang2025embodiedbench,liu2025robogptr1,xu2026roboagent}.
Recent MLLMs have demonstrated strong instruction understanding, visual perception, and planning capabilities in such environments
~\cite{ahn2022saycan,yao2023react,song2023llmplanner,being0}.
To further improve their reliability, a growing body of work directly post-trains the executor using expert or model-generated trajectories, environment-grounded supervision, and task-level interaction rewards,
typically through supervised fine-tuning, imitation learning, or reinforcement learning
~\cite{zhai2024rl4vlm,wang2025vagen,chen2025era,wu2026orbit,xu2026roboagent}.
These methods enable the model to internalize task- and environment-specific experience in its parameters. Nevertheless, challenging instructions and long-horizon interactions continue to expose failures. In such cases, effective adaptation requires more than generating a plan: the agent must reflect on its execution outcomes and preserve useful experience for future decisions.


\begin{figure}
    \centering
    \includegraphics[width=\linewidth]{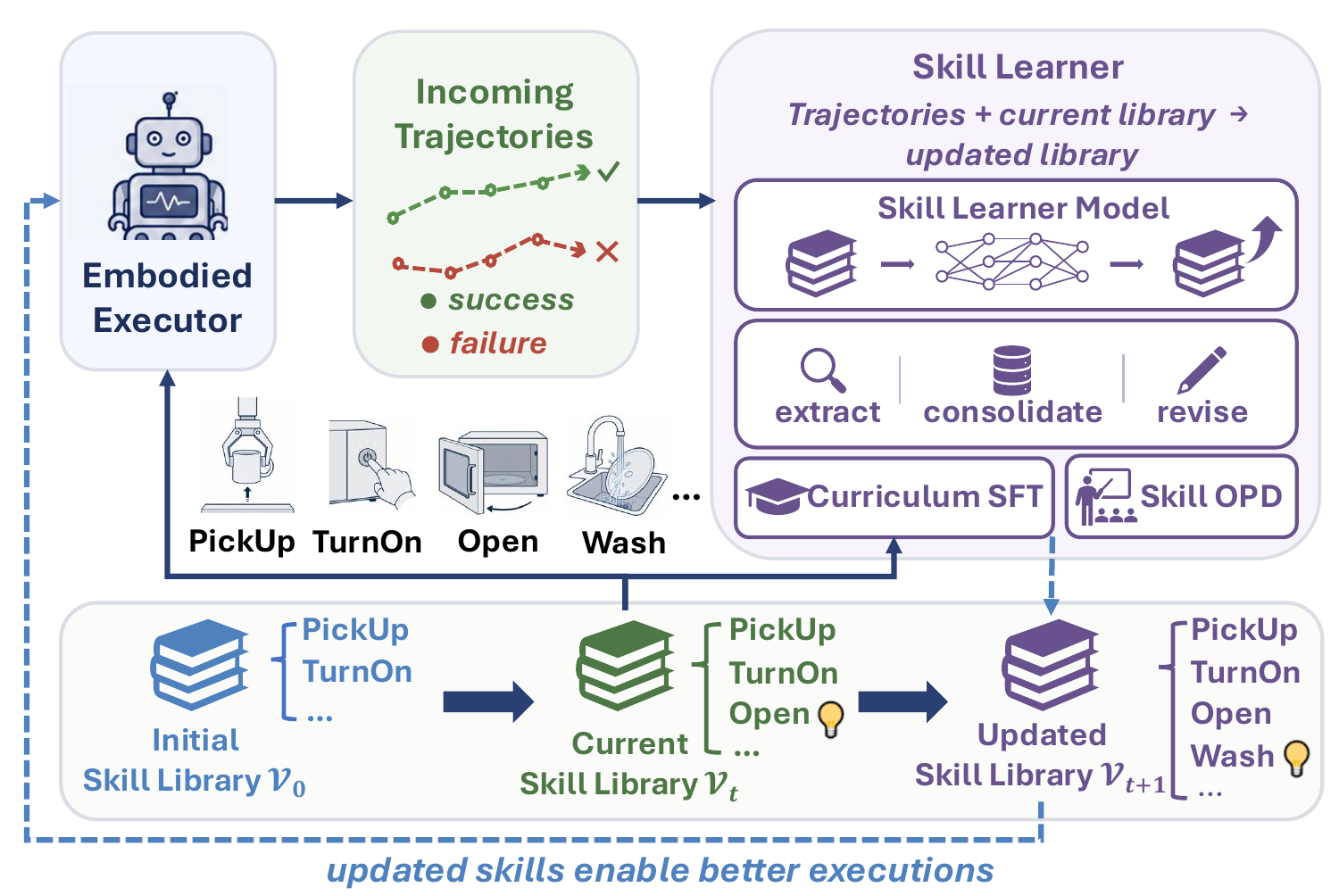}
    \caption{\textbf{Overview of our skill learning framework.}
A trainable skill learner extracts, consolidates, and revises reusable skills from successful and failed execution trajectories, progressively updating an external skill library. The updated skills guide the embodied executor toward improved task execution.}
    \label{fig:teaser}
\end{figure}

Therefore, a line of experience-augmented methods externalizes such experience as persistent memories or reusable skills. Memory-based methods extract and retrieve episodic or semantic records ~\cite{zhang2025gmemory,chhikara2025mem0,zhang2025memgen}, whereas skill-based methods distill trajectories into action procedures, lessons, or executable rules ~\cite{ni2026trace2skilldistilltrajectorylocallessons,jiang2026xskillcontinuallearningexperience, yang2026autoskillexperiencedrivenlifelonglearning, ju2026embodiskill,ding2026memcompiler}.
However, most of the current experience-augmented works rely on fixed, prompt-engineered workflows to generate, refine, merge, or remove skill entries. Consequently, although the library content evolves, its update policy does not learn from accumulated experience or downstream execution outcomes, limiting its adaptability across tasks and environments. We therefore ask: \emph{Can an embodied agent learn not only new skills, but also how to grow and refine those skills?}

To address this, we propose \textbf{\projname{}}, which treats the skill library as continuously evolving procedural knowledge and the skill learner as a learnable update policy. As shown in the Figure~\ref{fig:teaser}, the learner and the library are improved iteratively. The current library first guides task execution and shapes the collected trajectories, which are then used to improve the learner's skill-refinement policy. The updated learner, in turn, extracts reusable experience from these trajectories and produces refined skills. The updated library guides the next round of execution and provides new experience for further learning. In this way, the policy for refining skills and the procedural knowledge being refined improve together.


We instantiate this progression through three training stages. Stage~0 uses successful oracle trajectories to cold-start skill generation and library maintenance. 
Stage~1 contrasts successful and failed trajectories from heterogeneous executors on the same tasks, teaching the learner to identify invalid condition for skill application and derive strategies for failure recovery. 
Finally, Stage~2 introduces online skill-refinement policy distillation (\textsc{Skill OPD}), where the learner generates edits from incoming trajectories and aligns its behavior with a stronger teacher on its own edit distribution. Each stage inherits the learner and skill library from the preceding stage, progressively improving both the update policy and the procedural knowledge available to the executor.


We evaluate \projname{} on two high-level embodied task-planning benchmarks, EB-ALFRED and EB-Habitat, from
EmbodiedBench~\cite{yang2025embodiedbench}. Across heterogeneous executors, \projname{} consistently improves task performance and outperforms the strongest experience-augmented baselines by 9.7 and 2.6 percentage points, respectively. These findings suggest that sustained agent improvement depends not merely on evolving what an agent knows, but on learning how its procedural knowledge should evolve through experience. Our main contributions are:
\begin{itemize}
    \item We formulate skill evolution as a coupled learning problem and introduce a new framework that treats the skill library as evolving procedural knowledge and the skill learner as a learnable update policy.
    \item We propose a two-stage training framework combining curriculum supervised fine-tuning with online skill-edit distillation, enabling failure-aware skill maintenance and iterative refinement.
    \item We outperform the strongest experience-based baselines on EB-ALFRED and EB-Habitat while consistently improving multiple frozen executors.
\end{itemize}

\section{Related Works}

\subsection{Embodied Task Planning}

Embodied task planning requires agents to translate language instructions and visual observations into coherent action sequences. Recent work primarily improves this capability by post-training MLLM/VLM executors. RL4VLM applies reinforcement learning to VLM decision policies~\cite{zhai2024rl4vlm}, while REBP and RoboGPT-R1 combine supervised initialization with reinforcement learning for structured reasoning, task completion, and action consistency~\cite{wu2025reinforcedreasoning,liu2025robogptr1}. VAGEN reinforces state estimation and transition modeling under partial observability~\cite{wang2025vagen}. More recent methods exploit richer task supervision: ERA combines embodied priors with online RL~\cite{chen2025era}, ORBIT performs on-policy fine-tuning with offline rewards~\cite{wu2026orbit}, and RoboAgent trains coordinated vision-language capabilities through behavior cloning, trajectory aggregation, and reinforcement learning~\cite{xu2026roboagent}. These approaches improve planning by internalizing task knowledge and interaction policies into the executor's parameters.

\subsection{Experience-Augmented Agents}

Because executor post-training typically requires task-specific trajectories, another line of work externalizes experience as reusable skills or memories. Prompt-based methods summarize trajectories and maintain external knowledge through predefined operations: Trace2Skill consolidates trajectory-level lessons~\cite{ni2026trace2skilldistilltrajectorylocallessons}, AutoSkill maintains skill artifacts through addition, merging, and removal~\cite{yang2026autoskillexperiencedrivenlifelonglearning}. Memory-based methods organize experience into different memory structures~\cite{zhang2025gmemory,zhang2025memgen,ding2026memcompiler}, while ELITE and EmbodiSkill use reflective updates to refine reusable strategies or procedural skills~\cite{wei2026elite,ju2026embodiskill}. However, their maintenance logic is primarily prescribed through prompts and heuristics. Therefore, researches like MemCtrl trains a memory gate to retain or discard observations~\cite{dorbala2026memctrl}, SkillOS learns a curator for an external skill repository~\cite{ouyang2026skillos}. In contrast, \projname{} progressively trains a dedicated learner to grow and refine a persistent skill library.

\begin{figure*}
    \centering
    \includegraphics[width=\linewidth]{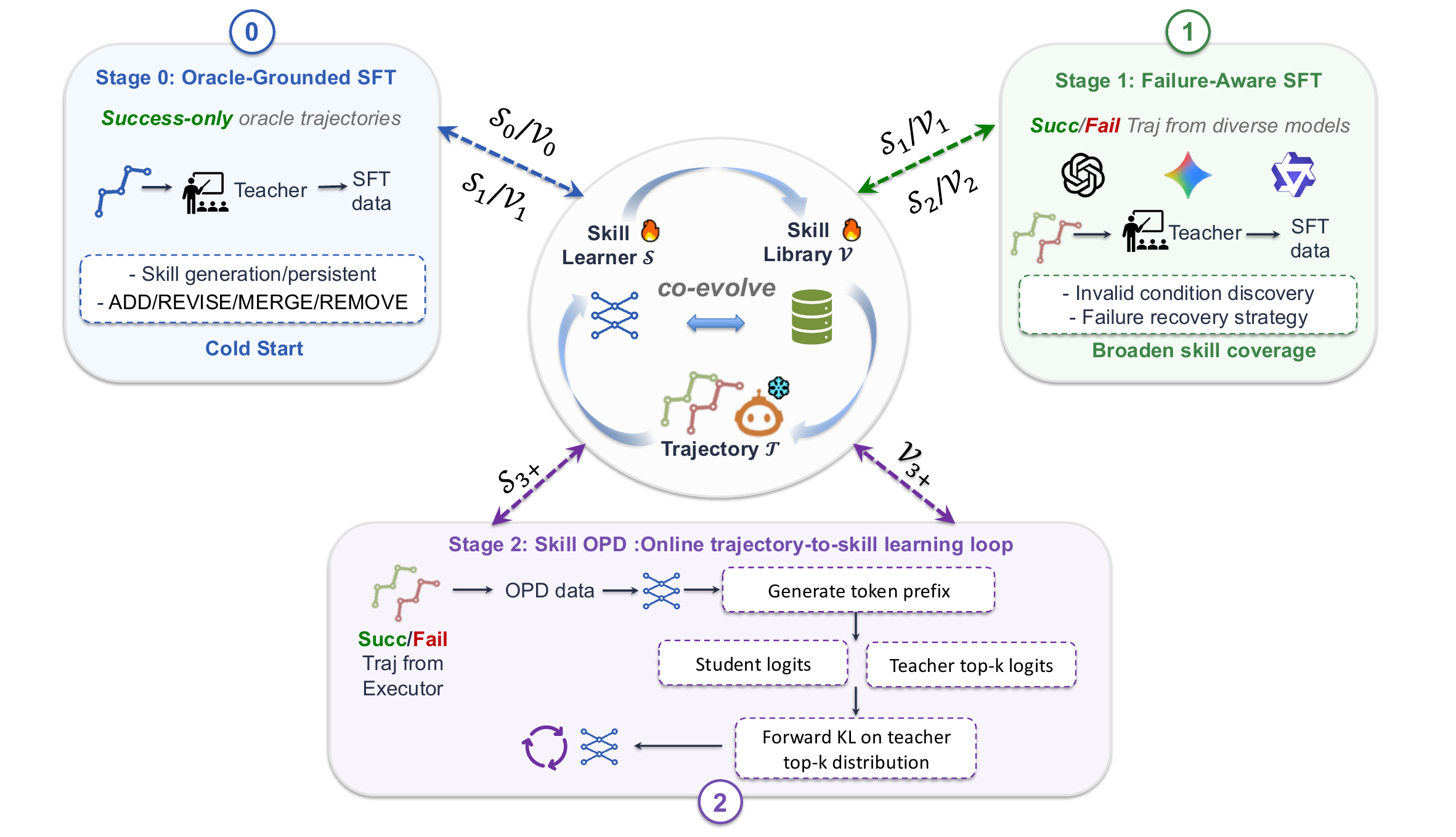}
    \caption{\textbf{Training pipeline of \projname{}.} Stages~0--1 use oracle-grounded and failure-aware SFT to initialize skill learning while Stage~2 applies \textsc{Skill OPD} to match a teacher's token distributions on student-generated prefixes via forward KL.}
    \label{fig:method_overview}
\end{figure*}


\section{Methodology}
\subsection{Problem Formulation}
Our agent system consists of an embodied task executor, a persistent
skill library, and a learnable skill learner: the library guides task
execution, while the learner converts the resulting trajectories into
library updates, forming a closed loop of skill-level self-evolution.

\paragraph{Embodied task executor.}
We consider an embodied task executor that interacts with a physically
grounded environment $\mathcal{E}$ under natural-language instructions.
Each task is modeled as a partially observable decision process. Given an
instruction $I$, the executor receives an observation $o_t$ at each time
step $t$ and selects an action $a_t$ from the action space $\mathcal{A}$.
We denote the executor policy by $\pi_{\mathrm{exec}}$. An execution
trajectory is represented as
\begin{equation}
    \tau =
    \left(
        I,o_1,a_1,\ldots,o_T,a_T,y
    \right),
\end{equation}
where $T$ is the trajectory length and $y\in\{0,1\}$ denotes the final task-success outcome. Throughout our framework, the parameters of $\pi_{\mathrm{exec}}$ remain frozen.


\paragraph{Skill Library.}
The skill library $\mathcal{V}$ stores reusable procedural knowledge
available to the executor for
embodied task execution. The skill library is a collection of skill cards:
\begin{equation}
    \mathcal{V}
    =
    \left\{
        v_j
    \right\}_{j=1}^{|\mathcal{V}|}.
\end{equation} We define a skill card as a parameterized procedural unit. Each skill card $v_j$ consists of an action pattern sequence $p_j$ and a structured usage specification $u_j$:
\begin{equation}
    v_j = (p_j,u_j),
    \qquad
    p_j =
    \left[
        (a_1,\phi_1),\ldots,(a_L,\phi_L)
    \right],
\end{equation}
where $a_l\in\mathcal{A}$ is a primitive action and $\phi_l$ denotes its
arguments, such as an object category or a target state. For example, the pattern
$(\texttt{pick\_up}, \texttt{object\_target})$
could describe a reusable ``picking-up'' skill. The usage specification $u_j$ describes the semantic scope and execution constraints of the skill, such as \emph{when} the skill is applicable and \emph{how} the executor should recover from failure.



\paragraph{Skill-guided execution.}
The task executor is augmented with the
current skill library $\mathcal{V}$. The library provides procedural
guidance for action planning, including applicable action patterns,
execution preconditions, expected effects, and recovery strategies. Let $h_t$ denotes the interaction history available at step $t$. The executor selects
its next action according to
\begin{equation}
    a_t
    \sim
    \pi_{\mathrm{exec}}
    \left(
        \cdot
        \mid
        I,h_t,\mathcal{V}
    \right).
    \label{eq:skill_guided_execution}
\end{equation}
%
%
Although the parameters of $\pi_{\mathrm{exec}}$ remain unchanged, different
versions of the skill library induce different action distributions and
therefore different trajectory distributions.

\paragraph{Skill learner.}
We introduce the skill learner
$\mathcal{S}$ as a parameterized library-update policy. It interprets execution trajectories, extract reusable procedural experience, and determine how the persistent skill library should grow and be refined over time. At update round $i$, the current library $\mathcal{V}_i$ is provided to
the task executor to collect a new trajectory set $\mathcal{T}_i$. Then the skill learner predicts a set of structured edits to grow the skill library like:
\begin{equation}
    \Delta =
    \mathcal{S}
    \left(
        \mathcal{V},\mathcal{T}
    \right).
\end{equation}
Here, $\Delta$ may contain operations that \texttt{ADD}, \texttt{REVISE}, \texttt{MERGE}, or \texttt{REMOVE}
skill cards. The predicted edits are applied to produce the next skill
library:
\begin{equation}
   \mathcal{V}
    =
    \operatorname{Apply}
    \left(
    \mathcal{V}, \Delta
    \right).
\end{equation}


The skill learner and skill library thus form a coupled evolution process. An improved skill learner produces more reliable library edits, while the updated library changes the executor's future trajectory distribution and provides new on-policy evidence for learning subsequent update policies.

\subsection{\projname{} Overview}
As illustrated in Figure~\ref{fig:method_overview}, instead of relying on
a fixed, prompt-engineered procedure for constructing and maintaining a
skill library, \projname{} establishes a coupled update cycle
between a learnable skill learner and a persistent skill library.

\textbf{(1) Learner update}
($\mathcal{S}_i \xrightarrow{(\mathcal{V}_i,\mathcal{T}_i)}
\mathcal{S}_{i+1}$).
At round $i$, the current library $\mathcal{V}_i$ guides the executor and
induces trajectories $\mathcal{T}_i$, which provide supervision for
improving the skill learner.

\textbf{(2) Library update}
($\mathcal{V}_i \xrightarrow{\mathcal{S}_{i+1}}
\mathcal{V}_{i+1}$).
The updated learner interprets $\mathcal{T}_i$ in the context of
$\mathcal{V}_i$ and produces structured edits to construct
$\mathcal{V}_{i+1}$.

The updated library subsequently changes the executor's trajectory
distribution and provides new interaction evidence for the next round:
\begin{equation}
    \mathcal{V}_i
    \rightarrow \mathcal{T}_i
    \rightarrow \mathcal{S}_{i+1}
    \rightarrow \mathcal{V}_{i+1}
    \rightarrow \mathcal{T}_{i+1}.
\end{equation}
We instantiate this coupled progression through three stages:
oracle-grounded SFT, failure-aware SFT, and online skill-refinement policy
distillation (\textsc{Skill OPD}). Each stage inherits the learner and
library from the preceding stage, progressively improving both the
skill-refinement policy and the procedural knowledge. We
introduce each stage below.

\subsubsection{Initial Skill Library Construction}
We construct the initial library through action-pattern discovery inspired by byte-pair encoding (BPE)~\cite{sennrich2016bpe}. 
We first collect successful trajectories from the executor without skill augmentation and convert them into sequences of primitive actions. 
Frequent adjacent action subsequences are then iteratively merged to discover recurring multi-step patterns.
Finally, the initial skill learner $\mathcal{S}_0$, initialized from a pretrained base model, consolidates these patterns into structured skill cards, forming the initial library $\mathcal{V}_0$.

\subsection{Curriculum Training of the Skill Learner}

Following curriculum learning, which organizes training examples in a meaningful progression from simpler to more difficult concepts~\cite{bengio2009curriculum}, we cold-start the skill learner through two stages.
Stage 0 uses successful oracle trajectories to establish fundamental skill generation, editing, and consolidation capabilities.
Stage 1 subsequently introduces heterogeneous executor rollouts containing both successes and failures, enabling the learner to perform failure-aware skill attribution and revision.


\paragraph{Structured skill-editing supervision.}

At stage $i$, the learner receives the current library
$\mathcal{V}_i$ and a trajectory batch $B_i\subseteq\mathcal{T}_i$. A strong
teacher model generates a structured target edit set:
\begin{equation}
    \Delta_i^{*}
    =
    \mathcal{G}
    \left(
        \mathcal{V}_i,B_i
    \right),
\end{equation}
where $\mathcal{G}$ denotes the teacher-based data generation process.
Depending on the trajectory evidence and current library state,
$\Delta_i^{*}$ may contain \texttt{ADD}, \texttt{REVISE}, \texttt{MERGE}, and \texttt{REMOVE} operations. Given the resulting curriculum dataset $\mathcal{D}_k$, the skill learner is optimized with the autoregressive SFT objective:
\begin{equation}
    \mathcal{L}_{\mathrm{SFT}}^{(i)}
    =
    -
    \mathbb{E}_{(x,y)\sim\mathcal{D}_i}
    \left[
        \sum_{t=1}^{|y|}
        \log
        p_{\mathcal{S}_{i+1}}
        \left(
            y_t \mid x,y_{<t}
        \right)
    \right],
\end{equation}
where $x$ contains the current library and trajectory evidence, and $y$
denotes the teacher-generated editing output.

\subsubsection{Stage 0: Oracle-Grounded Skill Editing}

Stage 0 uses only successful oracle trajectories. This controlled setting allows the learner to acquire basic skill-editing capabilities before it must reason about noisy and potentially ambiguous failures.

To simulate the streaming arrival of executor experience, we randomly sample $N$ oracle trajectories to form each trajectory batch:
\begin{equation}
    B_{0,b}
    \subseteq
    \mathcal{T}^{\mathrm{oracle}},
    \qquad
    |B_{0,b}|=N,
\end{equation}
where $b$ denotes the batch index. We construct three complementary types of Stage-0 supervision.

\paragraph{Skill generation from an empty library.}

The first type removes all existing skill context and asks the teacher to
construct reusable skill cards directly from a trajectory batch:
\begin{equation}
    \Delta_{0,b}^{\mathrm{gen}}
    =
    \mathcal{G}
    \left(
        \varnothing,B_{0,b}
    \right).
\end{equation}
These examples teach the learner to identify reusable procedural abstractions
without relying on pre-existing skill cards.

\paragraph{Editing the initialized library.}

The second type provides the BPE-initialized library $\mathcal{V}_0$ together
with the same trajectory evidence:
\begin{equation}
    \Delta_{0,b}^{\mathrm{edit}}
    =
    \mathcal{G}
    \left(
        \mathcal{V}_0,B_{0,b}
    \right).
\end{equation}
The teacher determines whether the evidence supports adding a missing skill,
revising an incomplete skill, merging overlapping cards, removing an
unsupported card, or preserving the existing library. These examples train
the learner to make localized edits rather than regenerate the whole library.

\paragraph{Cross-batch library consolidation.}

Edits obtained from individual batches may describe the same procedure at different granularities or introduce partially conflicting applicability conditions. We therefore construct an additional consolidation task over
$M$ independently sampled batches:
\begin{equation}
    \mathcal{C}_{0}
    =
    \mathcal{G}_{\mathrm{merge}}
    \left(
        \mathcal{V}_0,
        \left\{
            \Delta_{0,b}^{\mathrm{edit}}
        \right\}_{b=1}^{M}
    \right),
\end{equation}
where $\mathcal{C}_0$ is a globally consolidated editing target. The
consolidation process removes semantic duplicates, merges compatible action
patterns, reconciles overlapping trigger conditions, and preserves distinct
skills when their preconditions or effects differ. This task teaches the
learner to maintain library-level consistency beyond a single local update.

The complete Stage-0 dataset is
\begin{equation}
    \mathcal{D}_0
    =
    \mathcal{D}_0^{\mathrm{gen}}
    \cup
    \mathcal{D}_0^{\mathrm{edit}}
    \cup
    \mathcal{D}_0^{\mathrm{con}},
\end{equation}
where the three subsets correspond to skill generation, library-conditioned
editing, and cross-batch consolidation. Training on $\mathcal{D}_0$ updates
the initial learner $\mathcal{S}_0$ into $\mathcal{S}_1$. The trained learner
then evolves the initial library:
\begin{equation}
    \Delta_0 =
    \mathcal{S}_1
    \left(
        \mathcal{V}_0,\mathcal{T}_0
    \right),
    \qquad
    \mathcal{V}_1 =
    \operatorname{Apply}
    \left(
        \mathcal{V}_0,\Delta_0
    \right).
\end{equation}

\subsubsection{Stage 1: Failure-Aware Learning}

Stage 0 exposes the learner only to successful oracle behavior. However, an online skill learner must also determine whether a failed execution reveals a missing action, an incorrect action or an invalid environmental condition. Stage 1 introduces trajectories generated by a heterogeneous pool of executor models, covering both successful and failed task executions.

Instead of randomly mixing trajectories from unrelated tasks, we construct each batch from multiple executors attempting the same task. For task $q$, the corresponding comparison batch is
\begin{equation}
    B_{1,q}
    =
    \left\{
        \tau_{q,m}
        \mid
        m\in\mathcal{M}
    \right\},
\end{equation}
where $\mathcal{M}$ denotes the executor pool. Since all trajectories share
the same instruction and success criterion, their behavioral differences
provide controlled evidence for identifying generalizable skill content.


For each same-task batch, the teacher jointly analyzes all trajectories and
produces a failure-aware editing target:
\begin{equation}
    \Delta_{1,q}^{*}
    =
    \mathcal{G}
    \left(
        \mathcal{V}_1,B_{1,q}
    \right).
\end{equation}
These examples constitute the Stage-1 dataset $\mathcal{D}_1$. Fine-tuning
on $\mathcal{D}_1$ updates $\mathcal{S}_1$ into the failure-aware learner
$\mathcal{S}_2$, which subsequently produces the next library:
\begin{equation}
    \Delta_1 =
    \mathcal{S}_2
    \left(
        \mathcal{V}_1,\mathcal{T}_1
    \right),
    \qquad
    \mathcal{V}_2 =
    \operatorname{Apply}
    \left(
        \mathcal{V}_1,\Delta_1
    \right).
\end{equation}
In summary, Stage~0 teaches the learner to perform structured library edits, while Stage~1 teaches it to identify invalid conditions and recovery strategies by contrasting successful and failed trajectories.

\subsection{Skill OPD}
Although curriculum SFT provides a strong initialization, its supervision is
constructed from a fixed output distribution. We therefore introduce Skill OPD, adapting knowledge distillation~\cite{hinton2015distilling} and on-policy sequence distillation~\cite{agarwal2024onpolicy,sheng2024hybridflow,li2026rethinkingopd} to skill-library editing. Unlike offline SFT, Skill OPD trains the learner on contexts induced by its own generated outputs, reducing the train--inference distribution mismatch in autoregressive skill editing.


At OPD iteration $i$, we group incoming executor trajectories into editing batches. Conditioned on the current skill library $\mathcal{V}_i$ and a trajectory batch, the current learner $\mathcal{S}_i$ samples multi-turn editing rollouts from its own output distribution. Each rollout contains proposed edit operations together with the intermediate library states induced by applying and merging those operations.

A frozen stronger teacher computes token distributions on the same student-generated prefixes. Skill OPD does not use task rewards or policy-gradient updates; instead, it directly minimizes a top-$K$ approximation to the forward KL divergence from the teacher to the student. Let $c_\ell$ denote a student-visited skill-editing context at token position $\ell$, $p_{\mathrm{T}}^{(K)}$ the teacher distribution restricted and renormalized over its top-$K$ support, and $p_{\mathcal{S}_i}$ the student distribution. The OPD objective is
\begin{equation}
    \mathcal{L}_{\mathrm{OPD}}^{(i)}
    =
    \mathbb{E}_{c_\ell}
    \left[
        D_{\mathrm{KL}}
        \left(
            p_{\mathrm{T}}^{(K)}
            (\cdot\mid c_\ell)
            \,\Vert\,
            p_{\mathcal{S}_i}
            (\cdot\mid c_\ell)
        \right)
    \right].
\end{equation}
The expectation is taken over token contexts visited by the student's own editing rollouts. Minimizing this objective updates the learner from $\mathcal{S}_i$ to $\mathcal{S}_{i+1}$. The improved learner is then used to regenerate the edits and produce the next skill library.

The updated library guides the executor in the next iteration, producing a new trajectory distribution for further skill-learner optimization. Skill OPD therefore forms a recursive loop in which the learner improves the library, and the evolving library continually supplies new on-policy experience for
improving the learner.

\section{Experiments}

\subsection{Experimental Setup}

\textbf{Benchmarks.}
We evaluate on 2 high-level embodied task-planning benchmarks from EmbodiedBench~\cite{yang2025embodiedbench}: EB-ALFRED and EB-Habitat. Both of them contain 6 splits that assess distinct capability dimensions: basic capabilities (Base), commonsense reasoning (Com.), complex-instruction following (Comp.), visual recognition (Vis.), spatial reasoning (Spat.), and long-horizon planning (Long).

\paragraph{Implementation Details.}
We use Qwen3-VL-8B-Instruct~\cite{bai2025qwen3} as the base model and Qwen3.7-Max~\cite{alibabacloud2026qwen37max} to construct supervision. Skill edits are generated from batches of 10 trajectories, and each consolidation step merges 15 candidate edits. SFT is performed for three epochs with a learning rate of $5\times10^{-7}$. For Skill OPD, we use Qwen3-VL-32B-Instruct~\cite{bai2025qwen3} as the teacher and train for 20 epochs with a global batch size of 8 and one rollout per prompt. We apply top-$K$ forward-KL distillation with $K=32$ and no auxiliary
policy-gradient loss.

\paragraph{Baselines.}
We select baselines to distinguish three sources of performance improvement: backbone capability (\emph{No Skill}), direct executor post-training (\emph{Training-based MLLMs}), and the use of knowledge derived from experience (\emph{Experience-Augmented MLLMs}).


\subsection{Main Results}

\newcommand{\compactcite}[2]{(#1 et al.~\citeyear{#2})}
\begin{table*}[t]
\centering

\setlength{\tabcolsep}{3.2pt}
\renewcommand{\arraystretch}{1.08}
\resizebox{\textwidth}{!}{
\begin{tabular}{lccccccclcccccc}
\toprule
\textbf{Model / Method}
& \multicolumn{7}{c}{\textbf{EB-ALFRED}}
& \multicolumn{7}{c}{\textbf{EB-Habitat}} \\
\cmidrule(lr){2-8}
\cmidrule(lr){9-15}
& \textbf{Avg.}
& \textbf{Base}
& \textbf{Common}
& \textbf{Complex}
& \textbf{Visual}
& \textbf{Spatial}
& \textbf{Long}
& \textbf{Avg.}
& \textbf{Base}
& \textbf{Common}
& \textbf{Complex}
& \textbf{Visual}
& \textbf{Spatial}
& \textbf{Long} \\
\midrule

\multicolumn{15}{c}{\textit{No skill}} \\
\midrule


GPT-5.4
& 65.3 & 76 & 72 & 76 & 54 & 58 & 56
& 65.7 & 98 & 52 & 62 & 62 & 40 & 80 \\


Gemini-3-Flash
& 59.0 & 70 & 56 & 60 & 68 & 56 & 44 
& 51.0 & 86 & 52 & 46 & 62 & 30 & 30 \\




Qwen3.5-Flash
& 47.7 & 54 & 50 & 62 & 32 & 44 & 44
& 38.0 & 70 & 22 & 28 & 50 & 36 & 22 \\

Qwen3-VL-32B-Instruct
& 24.3 & 30 & 38 & 34 & 28 & 16 & 0
& 37.3 & 70 & 34 & 38 & 32 & 32 & 18 \\

\midrule
\multicolumn{15}{c}{\textit{Training-based MLLMs}} \\
\midrule

REBP~\cite{wu2025reinforcedreasoning}
& 35.6 & 54 & 42 & 46 & 28 & 38 & 6
& 20.0$^{*}$ & 56 & 8 & 18 & 16 & 14 & 8 \\




ORBIT$\dagger$~\cite{wu2026orbit}
& 72.4$\dagger$ & 86 & 62 & 82 & 72 & 60 & --
& 22.4$^{*}$ & 56 & 8 & 18 & 16 & 14 & -- \\

RoboGPT-R1~\cite{liu2025robogptr1}
& 55.3 & 62 & 56 & 64 & 50 & 50 & 50
& 22.0$^{*}$ & 64 & 8 & 18 & 20 & 12 & 10 \\

RoboAgent~\compactcite{Xu}{xu2026roboagent}
& 67.0 & 72 & 48 & 64 & 78 & 60 & 80
& 22.3$^{*}$ & -- & -- & -- & -- & -- & -- \\

ESCA~\cite{huang2026esca}
& 38.0 & 46 & -- & -- & -- & -- & 30
& 60.0 & 86 & -- & -- & -- & -- & 34 \\

ELITE~\cite{wei2026elite}
& 70.8 & 78 & 78 & 68 & 68 & 62 & --
& 67.0 & 90 & -- & -- & -- & -- & 44 \\

\midrule
\multicolumn{15}{c}{\textit{Experience-Augmented MLLMs (Executor: Qwen3-VL-32B)}} \\
\midrule

G-Mem~\cite{zhang2025gmemory}
& 25.3 & 34 & 36 & 38 & 24 & 14 & 6
& 44.0 & 88 & 24 & 36 & 50 & 32 & \underline{34} \\

Mem0~\cite{chhikara2025mem0}
& 30.3 & 40 & 42 & 40 & 26 & 26 & 8
& 38.3 & 84 & 24 & 26 & 42 & 34 & 20 \\

A-Mem~\cite{xu2025amem}
& 32.3 & 46 & 50 & 34 & 30 & 30 & 4
& 36.7 & 86 & 12 & 38 & 24 & 36 & 24 \\

LangMem~\cite{langchain2025langmem}
& 31.3 & 38 & 44 & 44 & 28 & 26 & 8
& 38.3 & 86 & 26 & 28 & 40 & 32 & 18 \\

MemGen~\compactcite{Zhang}{zhang2025memgen}
& 14.2 & 22 & 18 & 16 & 6 & 20 & 4
& 25.7 & 50 & 14 & 22 & 36 & 22 & 10 \\

EmbodiedSkill~\cite{ju2026embodiskill}
& -- & -- & -- & -- & -- & -- & --
& 52.3 & \textbf{96} & 44 & \underline{52} & 62 & 36 & 24 \\

MemCompiler~\cite{ding2026memcompiler}
& \underline{40.0} & \underline{52} & \textbf{54} & \underline{48} & \underline{38} & \underline{38} & \underline{10}
& \underline{55.7} & \underline{90} & \underline{58} & 44 & \underline{64} & \underline{40} & \textbf{38} \\



\textbf{\projname{} (Ours)}
& \textbf{49.7} & \textbf{66} & \underline{50} & \textbf{70} & \textbf{42} & \textbf{46} & \textbf{24}
& \textbf{58.3} & 82 & \textbf{68} & \textbf{60} & \textbf{76} & \textbf{42} & 22 \\

\midrule
\textbf{\textsc{\projname{}-GPT 5.4} (Ours)}
& \textit{67.7} & \textit{74} & \textit{74} & \textit{72} & \textit{60} & \textit{60} & \textbf{\textit{66}}
& \textbf{\textit{72.0}} & \textit{96} & \textbf{\textit{76}} & \textit{54} & \textbf{\textit{92}} & \textbf{\textit{54}} & \textit{60} \\

\bottomrule
\end{tabular}
}

\caption{
Success rate (\%) on EB-ALFRED and EB-Habitat. $^{\dagger}$ indicates that the average is computed over the five
reported splits, excluding the Long split. $^{*}$ denotes unseen/OOD evaluation. ``--'' denotes results that are not reported. The best results are shown in \textbf{bold}, and the second-best results are \underline{underlined}. \projname{} results with GPT-5.4 executor~\cite{openai2026gpt54} are \textit{italic} for distinction.
}
\label{tab:main_results}

\end{table*}

As shown in Table~\ref{tab:main_results}, \projname{} achieves
the best overall performance among experience-augmented methods under
the same Qwen3-VL-32B-Instruct executor.
It obtains average success rates of 49.7\% on EB-ALFRED and 58.3\% on EB-Habitat, exceeding the strongest prior experience-augmented baseline.
Among the reported split-level results, \projname{} performs best on five EB-ALFRED splits and four EB-Habitat splits.
The improvements are especially pronounced on Complex and Long in EB-ALFRED and on Complex and Visual in EB-Habitat, showing the benefit of converting experience into reusable guidance for complex instruction following and task breakdown.
\projname{} trails the strongest baselines on Base and Long in EB-Habitat, where highly variable target locations are less amenable to fixed action patterns, suggesting room for more adaptive search skills.

Some training-based methods, such as ELITE~\cite{wei2026elite}, achieve higher performance by post-training the executor on substantial task-specific data and interactions. In contrast, \projname{} trains a separate skill learner while keeping the executor unchanged. It can substantially improve the task performances by continually converting new trajectories into refined skills without repeatedly updating the executor's parameters.

Notably, \projname{}-augmented Qwen3-VL-32B-Instruct surpasses the
strong proprietary Qwen3.5-Flash, while \projname{}-augmented
GPT-5.4 achieves new state-of-the-art performance on EB-habitat benchmarks among all methods.
These results show that our learner--library co-evolution pipeline can strengthen heterogeneous executors, narrowing the gap to stronger foundation models with a compact skill learner.

\subsection{Ablation}
\paragraph{Generalization across task executors.}
We additional conduct a cross-executor generalization study. Following the multi-executor evaluation protocol of
MemCompiler~\cite{ding2026memcompiler}, we instantiate the task executor with GPT-5.2, Gemini-3-Flash~\cite{google2025gemini3flash} (Gemini for shot), and Qwen3-VL-32B-Instruct (Qwen), covering both proprietary and open-weight models. For each executor, we compare skill-augmented performance against its matched unaugmented baseline under the same evaluation protocol.

\begin{table}[t]
\centering

\setlength{\tabcolsep}{3.0pt}
\renewcommand{\arraystretch}{1.08}
\resizebox{0.9\columnwidth}{!}{
\begin{tabular}{llccc}
\toprule
\textbf{Method}
& \textbf{Executor}
& \textbf{No Aug.}
& \textbf{With Aug.}
& \textbf{$\Delta$} \\
\midrule


BrainMem
& Claude-3.5-Sonnet
& 64.0
& 74.7
& +10.7 \\

MemCtrl$^\ddagger$
& Gemma-3-12B
& 25.6
& 27.8
& +2.2 \\

MemCompiler
& Qwen3-VL-32B
& 19.0
& 40.0
& +21.0 \\

\midrule

\multirow{3}{*}{LangMem}
& GPT-5.2
& 32.0
& 46.7
& +14.7 \\

& Gemini-3-Flash
& 59.0
& \textbf{78.0}
& \textbf{+19.0} \\

& Qwen3-VL-32B
& 19.0
& 31.3
& +12.3 \\

\midrule

\multirow{3}{*}{\projname{} (Ours)}
& GPT-5.4
& 65.3
& 67.7
& +2.4 \\

& GPT-5.2
& 32.0
& \textbf{62.3}
& \textbf{+30.3} \\

& Gemini-3-Flash
& 59.0
& 74.3
& +15.3 \\

& Qwen3-VL-32B
& 24.3
& \textbf{49.7}
& \textbf{+25.4} \\

\bottomrule
\end{tabular}
}
\caption{
Average success rate on EB-ALFRED. $\Delta$ denotes the absolute improvement from the experience augmentation. $^{\ddagger}$ indicates that the average is computed over the five reported splits, excluding the Visual split.}
\label{tab:cross_executor_avg}
\end{table}

As shown in Table~\ref{tab:cross_executor_avg}, \projname{} consistently improves all three executors.
It increases the average success rate of GPT-5.4, GPT-5.2, Gemini and Qwen of 2.4, 30.3, 15.3, and 25.4 percentage points, respectively.
Among the experience-augmented methods shown with the same executor, \projname{} yields the largest reported matched improvement on GPT-5.2 and Qwen. 
These results demonstrate the potential of improving diverse base executors through an evolving external skill library and support our claim that \projname{} provides a general mechanism for sustained capability improvement without separately post-training each executor.


\paragraph{Effect of progressive skill-learner training.}
Table~\ref{tab:ablation_eb_alfred_sft} presents a stage-wise ablation of \projname{}'s training pipeline. The average success rate improves consistently across all subsequent training stages.
Stage~0 increases the average from 40.7\% to 42.3\% by teaching the learner basic skill generation and consolidation from successful
oracle trajectories. Introducing failure-aware skill editing in Stage~1 further raises performance to 45.3\%, with a particularly clear improvement on the Long split from 6\% to 14\%. Finally, Stage~2 achieves the largest incremental gain among the learned stages, improving the average success rate by 4.4 points. At the same time, the number of skill cards in this stage also grows to the greatest one with 29 cards.
These results indicate that during Skill OPD stage, skill learner effectively learn to extract more new and effective skills by aligning with the stronger teacher on the skill-edit distribution, reducing the discrepancy between offline supervision and the edits encountered during iterative library evolution.

\begin{table}[t]
\centering

\setlength{\tabcolsep}{3.0pt}
\renewcommand{\arraystretch}{1.08}
\resizebox{\columnwidth}{!}{
\begin{tabular}{l|c|ccccccc}
\toprule
\textbf{Setting}
& \textbf{\#Cards} & \textbf{Base} & \textbf{Com.}
& \textbf{Comp.} & \textbf{Vis.} & \textbf{Spat.}
& \textbf{Long} & \textbf{Avg.} \\
\midrule

No Skill
& 0
& 30 & 38 & 34 & 28 & 16 & 0 & 24.3 \\

Initial Skill (BPE)
& 17
& 60 & 36 & 50 & \textbf{46} & 42 & 10 & 40.7 \\

Stage 0 (SFT)
& 21
& 62 & 48 & 58 & 36 & 44 & 6 & 42.3 \\

Stage 1 (SFT)
& 17
& 66 & 46 & 60 & 40
& 46 & 14 & 45.3 \\

Stage 2 (OPD)
& \textbf{29} & \textbf{66} & \textbf{50}
& \textbf{70} & 42 & \textbf{46}
& \textbf{24} & \textbf{49.7} \\

\bottomrule
\end{tabular}
}
\caption{
Ablation of \projname{}'s progressive training pipeline on
EB-ALFRED. We report task success rate (SR, \%) across three training
stages: curriculum SFT (Stages 0--1) and online skill-edit distillation
(Stage 2).
\textbf{Executor}: Qwen3-VL-32B-Instruct;
\textbf{Student}: Qwen3-VL-8B-Instruct.
The best result in each column is shown in \textbf{bold}.
}
\label{tab:ablation_eb_alfred_sft}
\end{table}

\paragraph{Effect of failure-aware skill learning.}
To isolate the contribution of failure-aware supervision, we compare two
variants of Stage~1. The full variant (Failure-Aware) uses both successful and failed
trajectories, requiring the skill learner to contrast their action
sequences and environment feedback to identify invalid conditions and
derive corresponding recovery strategies. These findings are recorded in
the respective fields of each skill card. As a control, Stage~1
(Succ-Only) uses only successful trajectories and removes supervision for
these two failure-related fields while retaining the rest of the
skill-learning procedure.

As shown in Table~\ref{tab:failure_recovery_ablation}, simply adding more successful trajectories provides only a marginal improvement, increasing the average success rate from 42.3\% to 42.7\%.
In contrast, failure-aware skill learning raises the average to 45.3\%, outperforming the success-only variant by 2.6 percentage points and Stage~0 by 3.0 points.
The largest improvement appears on the Long split, where success increases from 8\% to 14\%. Failure-aware learning also improves Base and Complex by 4 points and Spatial by 2 points over the success-only control.
These results indicate that failed trajectories provide supervision that cannot be obtained merely by collecting additional successful examples: they expose invalid action patterns, violated preconditions, and corrective behaviors that help the executor recover during complex and long-horizon tasks.

\begin{figure}
    \centering
    \includegraphics[width=\linewidth]{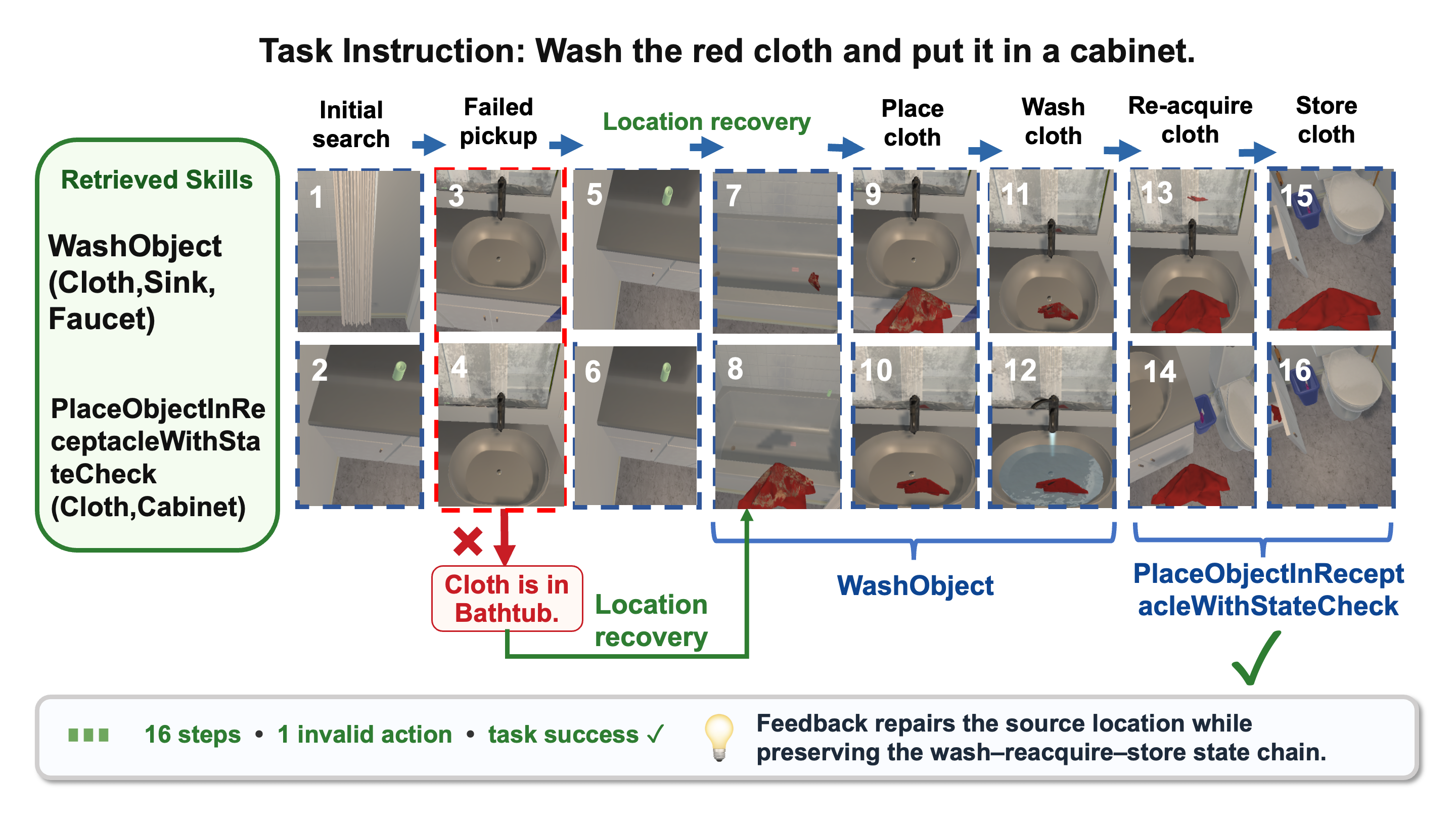}
    \caption{
    \textbf{A representative failure-recovery trajectory.}
    The \texttt{PickUp} failure triggered
    ``object not found'' strategy redirects the search to an alternative
    location. The executor then resumes the task and completes it using
    \texttt{WashObject} and \texttt{PlaceObjectReceptacle} in 16 steps.
    }
    \label{fig:failure_recovery_case}
\end{figure}

Figure~\ref{fig:failure_recovery_case} illustrates how failure-aware skills recovery from execution errors. After the initial pickup fails, the executor matches the environment feedback to an invalid condition recorded in the skill card and invokes the corresponding recovery action. This example demonstrates that learned recovery strategies make skill-guided execution more robust to unexpected
environment states.

\begin{table}[t]
\centering

\setlength{\tabcolsep}{3.2pt}
\renewcommand{\arraystretch}{1.08}
\resizebox{\columnwidth}{!}{
\begin{tabular}{lccccccc}
\toprule
\textbf{Variant}
& \textbf{Base}
& \textbf{Com.}
& \textbf{Comp.}
& \textbf{Vis.}
& \textbf{Spat.}
& \textbf{Long}
& \textbf{Avg.} \\
\midrule
Stage 0 
& 62 & \textbf{48} & 58 & 36 & 44 & 6 & 42.3 \\

Stage 1 (Succ-Only)
& 62 & 46 & 56 & 40 & 44 & 8 & 42.7 \\

Stage 1 (Failure-Aware)
& \textbf{66} & 46 & \textbf{60} & \textbf{40}
& \textbf{46} & \textbf{14} & \textbf{45.3} \\

\midrule
$\Delta$
& +4 & +0 & +4 & +0 & +2 & +6 & +2.6 \\
\bottomrule
\end{tabular}
}
\caption{
Ablation study of failure-aware recovery learning on EB-ALFRED.
All results are SR, \%.
$\Delta$ denotes the improvement of Stage~1 (Failure-Aware) over Stage~1 (Succ-Only).
}
\label{tab:failure_recovery_ablation}
\end{table}

\section{Conclusion}




This work reframes sustained improvement in embodied task planning as a problem of learning how interaction experience should be transformed into persistent procedural knowledge. 
\projname{} shows that reusable action patterns become substantially more effective when paired with explicit applicability conditions and failure-recovery strategies: together, they help executors decompose complex instructions, avoid repeated failure modes, and recover when execution deviates from the expected procedure. 
Rather than relying on a fixed, prompt-engineered maintenance workflow, \projname{} learns this transformation through a co-evolution of learner--library but with a frozen executor: the updated skill library shapes future trajectories, while those trajectories improve both the skill refinement policy and the knowledge retained in the next library. The consistent gains across training stages and heterogeneous executors validate the effectiveness and generality of this cycle. 
More broadly, our results suggest that learning to maintain an evolving procedural layer offers an effective path complementary to post-training the executor itself. Sustained capability growth can therefore arise not only from improving how an agent acts, but also from improving how it turns past experiences into reusable knowledge for future decisions.
\bibliography{aaai2027}

@misc{ni2026trace2skilldistilltrajectorylocallessons,
      title={Trace2Skill: Distill Trajectory-Local Lessons into Transferable Agent Skills}, 
      author={Jingwei Ni and Yihao Liu and Xinpeng Liu and Yutao Sun and Mengyu Zhou and Pengyu Cheng and Dexin Wang and Erchao Zhao and Xiaoxi Jiang and Guanjun Jiang},
      year={2026},
      eprint={2603.25158},
      archivePrefix={arXiv},
      primaryClass={cs.AI},
      url={https://arxiv.org/abs/2603.25158}, 
}

@inproceedings{jiang2026xskillcontinuallearningexperience,
  author    = "Jiang, Guanyu and Su, Zhaochen and Qu, Xiaoye and Fung, Yi R.",
  year      = 2026,
  title     = "{{XSkill}: Continual Learning from Experience and Skills in Multimodal Agents}",
  booktitle = "Proceedings of the 43rd International Conference on Machine Learning {(ICML)}",
}

@misc{yang2026autoskillexperiencedrivenlifelonglearning,
      title={AutoSkill: Experience-Driven Lifelong Learning via Skill Self-Evolution}, 
      author={Yutao Yang and Junsong Li and Qianjun Pan and Bihao Zhan and Yuxuan Cai and Lin Du and Jie Zhou and Kai Chen and Qin Chen and Xin Li and Bo Zhang and Liang He},
      year={2026},
      eprint={2603.01145},
      archivePrefix={arXiv},
      primaryClass={cs.AI},
      url={https://arxiv.org/abs/2603.01145}, 
}

@misc{chen2025era,
      title={ERA: Transforming VLMs into Embodied Agents via Embodied Prior Learning and Online Reinforcement Learning}, 
      author={Hanyang Chen and Mark Zhao and Rui Yang and Qinwei Ma and Ke Yang and Jiarui Yao and Kangrui Wang and Hao Bai and Zhenhailong Wang and Rui Pan and Mengchao Zhang and Jose Barreiros and Aykut Onol and ChengXiang Zhai and Heng Ji and Manling Li and Huan Zhang and Tong Zhang},
      year={2025},
      eprint={2510.12693},
      archivePrefix={arXiv},
      primaryClass={cs.AI},
      url={https://arxiv.org/abs/2510.12693}, 
}

@inproceedings{zhang2025gmemory,
  author    = "Zhang, Guibin and Fu, Muxin and Wang, Kun and Wan, Frank and Yu, Miao and Yan, Shuicheng",
  year      = 2025,
  title     = "{{G-Memory}: Tracing Hierarchical Memory for Multi-Agent Systems}",
  booktitle = "Advances in Neural Information Processing Systems {(NeurIPS)}",
  volume    = 38,
}

@inproceedings{xu2025amem,
  author    = "Xu, Wujiang and Liang, Zujie and Mei, Kai and Gao, Hang and Tan, Juntao and Zhang, Yongfeng",
  year      = 2025,
  title     = "{{A-MEM}: Agentic Memory for {LLM} Agents}",
  booktitle = "Advances in Neural Information Processing Systems {(NeurIPS)}",
  volume    = 38,
}

@article{chhikara2025mem0,
  title   = {Mem0: Building Production-Ready AI Agents with
             Scalable Long-Term Memory},
  author  = {Chhikara, Prateek and Khant, Dev and Aryan, Saket and
             Singh, Taranjeet and Yadav, Deshraj},
  journal = {arXiv preprint arXiv:2504.19413},
  year    = {2025}
}

@misc{langchain2025langmem,
  author       = "{LangChain}",
  year         = 2025,
  title        = "{{LangMem}}",
  howpublished = "\url{https://github.com/langchain-ai/langmem}",
  note         = "Accessed: 2026-05-04",
}

@inproceedings{zhang2025memgen,
  author    = "Zhang, Guibin and Fu, Muxin and Yan, Shuicheng",
  year      = 2026,
  title     = "{{MemGen}: Weaving Generative Latent Memory for Self-Evolving Agents}",
  booktitle = "International Conference on Learning Representations {(ICLR)}",
}

@article{ding2026memcompiler,
  title={MemCompiler: Compile, Don't Inject--State-Conditioned Memory for Embodied Agents},
  author={Ding, Xin and Wang, Xinrui and Yang, Yifan and Wu, Hao and Jiang, Shiqi and Zhang, Qianxi and Mi, Liang and Zhu, Hanxin and Li, Kun and Liu, Yunxin and others},
  journal={arXiv preprint arXiv:2605.07594},
  year={2026}
}

@article{ju2026embodiskill,
  title={EmbodiSkill: Skill-Aware Reflection for Self-Evolving Embodied Agents},
  author={Ju, Ruofei and Wang, Xinrui and Ding, Xin and Yang, Yifan and Wu, Hao and Jiang, Shiqi and Zhang, Qianxi and Wen, Hao and Li, Xiangyu and Wang, Weijun and others},
  journal={arXiv preprint arXiv:2605.10332},
  year={2026}
}

@inproceedings{yang2025embodiedbench,
  author    = "Yang, Rui and Chen, Hanyang and Zhang, Junyu and Zhao, Mark and Qian, Cheng and Wang, Kangrui and Wang, Qineng and Koripella, Teja Venkat and Movahedi, Marziyeh and Li, Manling and Ji, Heng and Zhang, Huan and Zhang, Tong",
  year      = 2025,
  title     = "{{EmbodiedBench}: Comprehensive Benchmarking Multi-modal Large Language Models for Vision-Driven Embodied Agents}",
  booktitle = "Proceedings of the 42nd International Conference on Machine Learning {(ICML)}",
  pages     = "70576--70631",
  volume    = 267,
  publisher = "PMLR",
}

@article{wei2026elite,
  title={ELITE: Experiential Learning and Intent-Aware Transfer for Self-improving Embodied Agents},
  author={Wei, Bingqing and Xia, Zhongyu and Liu, Dingai and Zhou, Xiaoyu and Lin, Zhiwei and Wang, Yongtao},
  journal={arXiv preprint arXiv:2603.24018},
  year={2026}
}

@article{dorbala2026memctrl,
  title={MemCtrl: Using MLLMs as Active Memory Controllers on Embodied Agents},
  author={Dorbala, Vishnu Sashank and Manocha, Dinesh},
  journal={arXiv preprint arXiv:2601.20831},
  year={2026}
}

@article{wu2025reinforcedreasoning,
  title   = {Reinforced Reasoning for Embodied Planning},
  author  = {Wu, Di and Fan, Jiaxin and Zang, Junzhe and Wang, Guanbo
             and Yin, Wei and Li, Wenhao and Jin, Bo},
  journal = {arXiv preprint arXiv:2505.22050},
  year    = {2025},
  url     = {https://arxiv.org/abs/2505.22050}
}

@inproceedings{wu2026orbit,
  author    = "Wu, Di and Fan, Jiaxin and Gu, Chloe and Wang, Guanbo and Yin, Wei and Li, Wenhao and Jin, Bo",
  year      = 2026,
  title     = "{On-Policy Reinforcement Fine-Tuning with Offline Reward for Multi-Step Embodied Planning}",
  booktitle = "Proceedings of the 64th Annual Meeting of the Association for Computational Linguistics {(ACL)}",
  pages     = "39277--39307",
  address   = "San Diego, California, United States",
  publisher = "Association for Computational Linguistics",
}

@article{liu2025robogptr1,
  title   = {{RoboGPT-R1}: Enhancing Robot Task Planning with
             Reinforcement Learning},
  author  = {Liu, Jinrui and Nie, Bingyan and Li, Boyu and Chen, Yaran
             and Wang, Yuze and He, Shunsen and Li, Haoran},
  journal = {arXiv preprint arXiv:2510.14828},
  year    = {2025},
  url     = {https://arxiv.org/abs/2510.14828}
}

@inproceedings{xu2026roboagent,
  author    = "Xu, Peiran and Zheng, Jiaqi and Mu, Yadong",
  year      = 2026,
  title     = "{{RoboAgent}: Chaining Basic Capabilities for Embodied Task Planning}",
  booktitle = "Proceedings of the IEEE/CVF Conference on Computer Vision and Pattern Recognition {(CVPR)}",
  pages     = "15276--15290",
}

@inproceedings{zhai2024rl4vlm,
  author    = "Zhai, Yuexiang and Bai, Hao and Lin, Zipeng and Pan, Jiayi and Tong, Shengbang and Zhou, Yifei and Suhr, Alane and Xie, Saining and LeCun, Yann and Ma, Yi and Levine, Sergey",
  year      = 2024,
  title     = "{Fine-Tuning Large Vision-Language Models as Decision-Making Agents via Reinforcement Learning}",
  booktitle = "Advances in Neural Information Processing Systems {(NeurIPS)}",
  volume    = 37,
  pages     = "110935--110971",
}

@inproceedings{wang2025vagen,
  author    = "Wang, Kangrui and Zhang, Pingyue and Wang, Zihan and Gao, Yaning and Li, Linjie and Wang, Qineng and Chen, Hanyang and Lu, Yiping and Yang, Zhengyuan and Wang, Lijuan and Krishna, Ranjay and Wu, Jiajun and Li, Fei-Fei and Choi, Yejin and Li, Manling",
  year      = 2025,
  title     = "{{VAGEN}: Reinforcing World Model Reasoning for Multi-Turn {VLM} Agents}",
  booktitle = "Advances in Neural Information Processing Systems {(NeurIPS)}",
  volume    = 38,
}

@inproceedings{huang2026esca,
  author    = "Huang, Jiani and Sethi, Amish and Kuo, Matthew and Keoliya, Mayank and Velingker, Neelay and Jung, JungHo and Lim, Ser Nam and Li, Ziyang and Naik, Mayur",
  year      = 2025,
  title     = "{{ESCA}: Contextualizing Embodied Agents via Scene-Graph Generation}",
  booktitle = "Advances in Neural Information Processing Systems {(NeurIPS)}",
  volume    = 38,
  pages     = "2828--2870",
}

@article{bai2025qwen3,
  title={{Qwen3-VL} Technical Report},
  author={Bai, Shuai and Cai, Yuxuan and Chen, Ruizhe and Chen, Keqin and Chen, Xionghui and Cheng, Zesen and Deng, Lianghao and Ding, Wei and Gao, Chang and Ge, Chunjiang and others},
  journal={arXiv preprint arXiv:2511.21631},
  year={2025}
}

@misc{alibabacloud2026qwen37max,
  author       = "{Alibaba Cloud}",
  year         = 2026,
  title        = "{{Qwen3.7-Max}}",
  howpublished = "\url{https://www.alibabacloud.com/help/en/model-studio/text-generation-model/}",
  note         = "Alibaba Cloud Model Studio documentation. Accessed: 2026-07-29",
}

@inproceedings{yao2023react,
  title     = {{ReAct}: Synergizing Reasoning and Acting in Language Models},
  author    = {Yao, Shunyu and Zhao, Jeffrey and Yu, Dian and Du, Nan and Shafran, Izhak and Narasimhan, Karthik and Cao, Yuan},
  booktitle = {International Conference on Learning Representations},
  year      = {2023},
  eprint    = {2210.03629},
  archivePrefix = {arXiv},
  url       = {https://arxiv.org/abs/2210.03629}
}

@article{ahn2022saycan,
  title   = {Do As I Can, Not As I Say: Grounding Language in Robotic Affordances},
  author  = {Ahn, Michael and Brohan, Anthony and Brown, Noah and Chebotar, Yevgen and Cortes, Omar and David, Byron and Finn, Chelsea and Fu, Chuyuan and Gopalakrishnan, Keerthana and Hausman, Karol and Herzog, Alex and Ho, Daniel and Hsu, Jasmine and Ibarz, Julian and Ichter, Brian and Irpan, Alex and Jang, Eric and Ruano, Rosario Jauregui and Jeffrey, Kyle and Jesmonth, Sally and Joshi, Nikhil J. and Julian, Ryan and Kalashnikov, Dmitry and Kuang, Yuheng and Lee, Kuang-Huei and Levine, Sergey and Lu, Yao and Luu, Linda and Parada, Carolina and Pastor, Peter and Quiambao, Jornell and Rao, Kanishka and Rettinghouse, Jarek and Reyes, Diego and Sermanet, Pierre and Sievers, Nicolas and Tan, Clayton and Toshev, Alexander and Vanhoucke, Vincent and Xia, Fei and Xiao, Ted and Xu, Peng and Xu, Sichun and Yan, Mengyuan and Zeng, Andy},
  journal = {arXiv preprint arXiv:2204.01691},
  year    = {2022},
  eprint  = {2204.01691},
  archivePrefix = {arXiv},
  url     = {https://arxiv.org/abs/2204.01691}
}

@inproceedings{song2023llmplanner,
  title     = {{LLM-Planner}: Few-Shot Grounded Planning for Embodied Agents with Large Language Models},
  author    = {Song, Chan Hee and Wu, Jiaman and Washington, Clayton and Sadler, Brian M. and Chao, Wei-Lun and Su, Yu},
  booktitle = {Proceedings of the IEEE/CVF International Conference on Computer Vision},
  pages     = {2998--3009},
  year      = {2023},
  eprint    = {2212.04088},
  archivePrefix = {arXiv},
  url       = {https://arxiv.org/abs/2212.04088}
}

@article{being0,
  title   = {Being-0: A Humanoid Robotic Agent with Vision-Language Models and Modular Skills},
  author  = {Haoqi Yuan and Yu Bai and Yuhui Fu and Bohan Zhou and Yicheng Feng and Xinrun Xu and Yi Zhan and Börje F. Karlsson and Zongqing Lu},
  journal = {arXiv preprint arXiv:2503.12533},
  year    = {2025},
  eprint  = {arXiv:2503.12533},
  archivePrefix = {arXiv},
  url     = {https://arxiv.org/abs/arXiv:2503.12533}
}

@inproceedings{bengio2009curriculum,
  title     = {Curriculum Learning},
  author    = {Bengio, Yoshua and Louradour, J\'{e}r\^{o}me and Collobert, Ronan and Weston, Jason},
  booktitle = {Proceedings of the 26th Annual International Conference on Machine Learning},
  pages     = {41--48},
  year      = {2009},
  doi       = {10.1145/1553374.1553380}
}

@article{hinton2015distilling,
  title   = {Distilling the Knowledge in a Neural Network},
  author  = {Hinton, Geoffrey and Vinyals, Oriol and Dean, Jeff},
  journal = {arXiv preprint arXiv:1503.02531},
  year    = {2015},
  eprint  = {1503.02531},
  archivePrefix = {arXiv},
  url     = {https://arxiv.org/abs/1503.02531}
}

@inproceedings{agarwal2024onpolicy,
  author    = "Agarwal, Rishabh and Vieillard, Nino and Zhou, Yongchao and Stanczyk, Piotr and Ramos Garea, Sabela and Geist, Matthieu and Bachem, Olivier",
  year      = 2024,
  title     = "{On-Policy Distillation of Language Models: Learning from Self-Generated Mistakes}",
  booktitle = "International Conference on Learning Representations {(ICLR)}",
}

@article{li2026rethinkingopd,
  title   = {Rethinking On-Policy Distillation of Large Language Models: Phenomenology, Mechanism, and Recipe},
  author  = {Li, Yaxuan and Zuo, Yuxin and He, Bingxiang and Zhang, Jinqian and Xiao, Chaojun and Qian, Cheng and Yu, Tianyu and Gao, Huan-ang and Yang, Wenkai and Liu, Zhiyuan and Ding, Ning},
  journal = {arXiv preprint arXiv:2604.13016},
  year    = {2026},
  eprint  = {2604.13016},
  archivePrefix = {arXiv},
  url     = {https://arxiv.org/abs/2604.13016}
}

@inproceedings{sennrich2016bpe,
  title     = {Neural Machine Translation of Rare Words with Subword Units},
  author    = {Sennrich, Rico and Haddow, Barry and Birch, Alexandra},
  booktitle = {Proceedings of the 54th Annual Meeting of the Association for Computational Linguistics},
  pages     = {1715--1725},
  year      = {2016},
  doi       = {10.18653/v1/P16-1162},
  eprint    = {1508.07909},
  archivePrefix = {arXiv},
  url       = {https://arxiv.org/abs/1508.07909}
}

@inproceedings{sheng2024hybridflow,
  author    = "Sheng, Guangming and Zhang, Chi and Ye, Zilingfeng and Wu, Xibin and Zhang, Wang and Zhang, Ru and Peng, Yanghua and Lin, Haibin and Wu, Chuan",
  year      = 2025,
  title     = "{{HybridFlow}: A Flexible and Efficient {RLHF} Framework}",
  booktitle = "Proceedings of the Twentieth European Conference on Computer Systems {(EuroSys)}",
  pages     = "1279--1297",
  publisher = "ACM",
}

@misc{google2025gemini3flash,
  author       = "{Google}",
  year         = 2025,
  title        = "{{Gemini 3 Flash Preview}}",
  howpublished = "\url{https://ai.google.dev/gemini-api/docs/models/gemini-3-flash-preview}",
  note         = "Accessed: 2026-07-29",
}

@article{comanici2025gemini25,
  title={{Gemini 2.5}: Pushing the Frontier with Advanced Reasoning,
         Multimodality, Long Context, and Next Generation Agentic
         Capabilities},
  author={Comanici, Gheorghe and Bieber, Eric and Schaekermann, Mike
          and others},
  journal={arXiv preprint arXiv:2507.06261},
  year={2025}
}

@misc{googledeepmind2025gemini3flash,
  title={{Gemini 3 Flash} Model Card},
  author={{Google DeepMind}},
  year={2025},
  howpublished={\url{https://deepmind.google/models/model-cards/gemini-3-flash/}},
  note={Published December 17, 2025}
}

@misc{googledeepmind2026gemini31pro,
  title={{Gemini 3.1 Pro} Model Card},
  author={{Google DeepMind}},
  year={2026},
  howpublished={\url{https://deepmind.google/models/model-cards/gemini-3-1-pro/}},
  note={Published February 19, 2026}
}

@misc{googledeepmind2026gemini35flash,
  title={{Gemini 3.5 Flash} Model Card},
  author={{Google DeepMind}},
  year={2026},
  howpublished={\url{https://deepmind.google/models/model-cards/gemini-3-5-flash/}},
  note={Published May 19, 2026}
}

@misc{openai2026gpt54,
  title={Introducing {GPT-5.4}},
  author={{OpenAI}},
  year={2026},
  howpublished={\url{https://openai.com/index/introducing-gpt-5-4/}},
  note={Published March 5, 2026}
}

@misc{alibabacloud2026qwen35flash,
  title={{Qwen3.5-Flash}},
  author={{Alibaba Cloud}},
  year={2026},
  howpublished={\url{https://www.alibabacloud.com/help/en/model-studio/vision-model/}},
  note={Model snapshot qwen3.5-flash-2026-02-23; accessed July 30, 2026}
}

@article{ouyang2026skillos,
  title={{SkillOS}: Learning skill curation for self-evolving agents},
  author={Ouyang, Siru and Yan, Jun and Chen, Yanfei and Han, Rujun and Wang, Zifeng and Mishra, Bhavana Dalvi and Meng, Rui and Li, Chun-Liang and Jiao, Yizhu and Zha, Kaiwen and others},
  journal={arXiv preprint arXiv:2605.06614},
  year={2026}
}

\appendix

\section{Additional Method Details}
\subsection{Training Data Construction}
\label{sec:sft_data}

Our curriculum supervised fine-tuning (SFT) consists of two stages. Stage~0 uses only successful oracle trajectories to establish the learner's basic skill-generation, skill-edit and skill-consolidate capabilities. For generation (Stage~0A), the $\texttt{current\_skill\_pool}$ is set to an empty list. We randomly sample $N=10$ trajectories as a batch from 300 oracle trajectories in both EB-ALFRED and EB-Habitat to construct supervision from \textit{Qwen-3.7-max}~\cite{bai2025qwen3}. Therefore we get several generated skill candidates for each batch, after that, we randomly sample $M=15$ candidates to construct skill-consolidate data. For editing (Stage~0B), the $\texttt{current\_skill\_pool}$ is the initial skill library $V_0$. Similarly as Stage~0A, we generate skill editing data from randomly trajectories batches from oracle.

Stage~1 groups trajectories produced by 8 heterogeneous executors according to the task: \textit{Gemini-2.5-Flash}~\cite{comanici2025gemini25}, \textit{Gemini-3-Flash}~\cite{googledeepmind2025gemini3flash}, \textit{Gemini-3.1-Pro}~\cite{googledeepmind2026gemini31pro}, \textit{Gemini-3.5-Flash}~\cite{googledeepmind2026gemini35flash}, \textit{GPT-5.4}~\cite{openai2026gpt54}, \textit{Qwen3.5-Flash}~\cite{alibabacloud2026qwen35flash}, and \textit{Qwen3-VL-8B/32B-Instruct}~\cite{bai2025qwen3}. By contrasting successful and failed executions of the same task, the learner generates failure-aware edits that capture additional \texttt{invalid\_conditions} and \texttt{recovery\_strategies}. Then from each group, we collect the corresponding skill editing with the skill pool of $V_1$, and consolidation data after that.

In Table~\ref{tab:sft_data_statistics}, skill-generation and local library-editing examples are jointly counted as \emph{edit samples}, while cross-batch consolidation examples are counted as \emph{merge samples}. 

\begin{table}[!t]
\centering

\setlength{\tabcolsep}{2.2pt}
\renewcommand{\arraystretch}{1.08}
\resizebox{\columnwidth}{!}{
\begin{tabular}{llrrrrr}
\toprule
\textbf{Benchmark}
& \textbf{Stage}
& \textbf{Succ.}
& \textbf{Fail.}
& \textbf{\# Gen/Edit}
& \textbf{\# Conso.}
& \textbf{Total} \\
\midrule

EB-ALFRED
& Stage 0A
& 300 & 0
& 1,000 & 400 & 1,400 \\

& Stage 0B
& 300 & 0
& 400 & 0 & 400 \\

& Stage 1
& 1,285 & 1,123
& 300 & 150 & 450 \\

\cmidrule(l){2-7}
& \textbf{Subtotal}
& \textbf{1,885} & \textbf{1,123}
& \textbf{1,700} & \textbf{550}
& \textbf{2,250} \\

\midrule

EB-Habitat
& Stage 0A
& 300 & 0
& 1,000 & 400 & 1,400 \\

& Stage 0B
& 150 & 0
& 300 & 100 & 400 \\

& Stage 1
& 1,316 & 1,084
& 300 & 150 & 450 \\

\cmidrule(l){2-7}
& \textbf{Subtotal}
& \textbf{1,766} & \textbf{1,084}
& \textbf{1,700} & \textbf{550}
& \textbf{2,250} \\

\bottomrule
\end{tabular}
}
\caption{
Statistics of the final curriculum SFT datasets.
Succ. and Fail. count source trajectories before batching, whereas Edit, Consolidation (Conso.), and Total count the resulting SFT samples.
}
\label{tab:sft_data_statistics}
\end{table}

\subsection{Training Algorithm}

\begin{algorithm}[!t]
\caption{
Progressive Learner--Library Co-Evolution in \projname{}.
}
\label{alg:Ex2Ex}
\begin{algorithmic}[1]

\REQUIRE Frozen task executor $\pi_{\mathrm{exec}}$;
initial skill learner $\mathcal{S}_0$;
oracle trajectories $\mathcal{T}^{\mathrm{ora}}$;
heterogeneous executors $\Pi_{\mathrm{het}}$;
SFT supervision teacher $P_{\mathrm{sup}}$;
OPD teacher $P_{\mathrm{T}}$;
OPD rounds $R$

\ENSURE Trained skill learner $\mathcal{S}_{R+1}$
and evolved skill library $\mathcal{V}_{R+1}$

\STATE \textbf{// Initial skill-library construction}

\STATE $\mathcal{P}_0 \leftarrow$
\textsc{BPE-PatternDiscovery}$(\mathcal{T}^{\mathrm{ora}})$

\STATE $\mathcal{V}_0 \leftarrow$
\textsc{InitializeSkillLibrary}$
(\mathcal{S}_0,\mathcal{P}_0,\mathcal{T}^{\mathrm{ora}})$

\STATE \textbf{// Stage 0: Oracle-Grounded SFT}

\STATE $\mathcal{D}_0 \leftarrow$
\textsc{ConstructOracleSupervision}$
(P_{\mathrm{sup}},\mathcal{V}_0,\mathcal{T}^{\mathrm{ora}})$

\STATE $\mathcal{S}_1 \leftarrow$
\textsc{SupervisedFineTune}$
(\mathcal{S}_0,\mathcal{D}_0)$
\COMMENT{Learner update}

\STATE $\mathcal{T}_0 \leftarrow$
\textsc{Rollout}$
(\pi_{\mathrm{exec}},\mathcal{V}_0)$

\STATE $\mathcal{V}_1 \leftarrow$
\textsc{UpdateLibrary}$
(\mathcal{S}_1,\mathcal{V}_0,\mathcal{T}_0)$
\COMMENT{Library update}

\STATE \textbf{// Stage 1: Failure-Aware SFT}

\STATE $\mathcal{T}_1 \leftarrow$
\textsc{CollectSameTaskRollouts}$
(\Pi_{\mathrm{het}},\mathcal{V}_1)$

\STATE $\mathcal{B}_1 \leftarrow$
\textsc{GroupByTask}$(\mathcal{T}_1)$

\STATE $\mathcal{D}_1 \leftarrow$
\textsc{ConstructFailureAwareSupervision}$
(P_{\mathrm{sup}},\mathcal{V}_1,\mathcal{B}_1)$

\STATE $\mathcal{S}_2 \leftarrow$
\textsc{SupervisedFineTune}$
(\mathcal{S}_1,\mathcal{D}_1)$
\COMMENT{Learner update}

\STATE $\mathcal{V}_2 \leftarrow$
\textsc{UpdateLibrary}$
(\mathcal{S}_2,\mathcal{V}_1,\mathcal{T}_1)$
\COMMENT{Library update}

\STATE \textbf{// Stage 2: On-Policy Distillation}

\FOR{$i=2,\ldots,R$}

    \STATE $\mathcal{T}_i \leftarrow$
    \textsc{Rollout}$
    (\pi_{\mathrm{exec}},\mathcal{V}_i)$

    \STATE $\mathcal{D}^{\mathrm{OPD}}_i \leftarrow$
    \textsc{ConstructOPDPrompts}$
    (\mathcal{V}_i,\mathcal{T}_i)$

    \STATE $\mathcal{S}_{i+1} \leftarrow$
    \textsc{OnPolicyDistillation}$
    (\mathcal{S}_i,P_{\mathrm{T}},
    \mathcal{D}^{\mathrm{OPD}}_i)$
    \COMMENT{Algorithm~\ref{alg:skill_opd}}

    \STATE $\mathcal{V}_{i+1} \leftarrow$
    \textsc{UpdateLibrary}$
    (\mathcal{S}_{i+1},\mathcal{V}_i,\mathcal{T}_i)$
    \COMMENT{Library update}

\ENDFOR

\RETURN $\mathcal{S}_{R+1},\mathcal{V}_{R+1}$

\STATE \textbf{// Subroutine: UpdateLibrary}$(\mathcal{S},\mathcal{V},\mathcal{T})$

\STATE Partition trajectories into batches
$\{\mathcal{B}_m\}_{m=1}^{M}$

\FOR{$m=1,\ldots,M$}

    \STATE $\Delta_m \leftarrow
    \mathcal{S}(\mathcal{V},\mathcal{B}_m)$

    \STATE \hspace{1em}
    $\Delta_m \in
    \{\texttt{ADD},\texttt{REVISE},
    \texttt{MERGE},\texttt{REMOVE}\}$

\ENDFOR

\STATE $\widetilde{\mathcal{V}} \leftarrow$
\textsc{ApplyEdits}$
(\mathcal{V},\bigcup_{m=1}^{M}\Delta_m)$

\STATE $\mathcal{V}' \leftarrow$
\textsc{HierarchicalConsolidation}$
(\widetilde{\mathcal{V}})$

\RETURN $\mathcal{V}'$

\end{algorithmic}
\end{algorithm}

Algorithm~\ref{alg:Ex2Ex} presents the pseudocode for the overall \projname{} framework. Specifically, we implement the Stage~2 training using the on-policy distillation (OPD) pipeline in \texttt{verl}. The training data are prompt-only: given a trajectory
prompt, the current student first generates a skill update from its own policy.
A frozen teacher then evaluates the student-generated sequence and provides its
Top-$K$ token distribution at each response position. We directly optimize the
student using a Top-$K$ approximation of the forward KL divergence. No task
reward or policy-gradient objective is used.

Let $\pi_\theta$ denote the student skill learner and $\pi_T$ the frozen teacher.
For a prompt $x$, the student first samples
\begin{equation}
    y \sim \pi_\theta(\cdot \mid x),
\end{equation}
where $y=(y_1,\ldots,y_T)$ is the student-generated skill update. At each
response position $t$, the teacher is evaluated on the student-induced context
$s_t=(x,y_{<t})$ and returns its Top-$K$ token set
\begin{equation}
    \mathcal{K}_t
    =
    \operatorname{TopK}_{K}
    \left(\pi_T(\cdot \mid s_t)\right).
\end{equation}

We minimize the truncated forward-KL objective
\begin{equation}
\label{eq:opd_topk_kl}
\begin{aligned}
\ell_t
&=
\sum_{v \in \mathcal{K}_t}
\pi_T(v \mid s_t)
\\
&\quad\times
\left[
\log \pi_T(v \mid s_t)
-
\log \pi_\theta(v \mid s_t)
\right].
\end{aligned}
\end{equation}
The loss is aggregated only over valid student-response tokens:
\begin{equation}
\label{eq:opd_loss}
    \mathcal{L}_{\mathrm{OPD}}
    =
    \frac{
        \sum_{t=1}^{T} m_t\,\ell_t
    }{
        \sum_{t=1}^{T} m_t
    },
\end{equation}
where $m_t$ is the response-token mask. In our implementation, $K=32$.
The student is updated by direct backpropagation through
$\mathcal{L}_{\mathrm{OPD}}$, while the teacher remains frozen.

\begin{algorithm}[t]
\caption{On-Policy Distillation for the Skill Learner}
\label{alg:skill_opd}
\begin{algorithmic}[1]

\REQUIRE Prompt-only dataset $\mathcal{D}$;
student $\pi_\theta$; frozen teacher $\pi_T$;
Top-$K$ size $K$; chunk size $C$
\ENSURE Distilled student $\pi_\theta$

\FOR{each training step}

    \STATE Sample prompt batch
    $\mathcal{B}=\{x_i\}_{i=1}^{B}$ from $\mathcal{D}$

    \STATE \textbf{// On-policy student rollout}
    \FOR{each $x_i \in \mathcal{B}$}
        \STATE Sample
        $y_i \sim \pi_\theta(\cdot \mid x_i)$
        \STATE Form sequence $z_i=[x_i;y_i]$
        \STATE Construct response mask $m_i$
    \ENDFOR

    \STATE \textbf{// Frozen-teacher supervision}
    \FOR{each student-generated sequence $z_i$}
        \STATE Evaluate frozen $\pi_T$ on $z_i$
        \FOR{each response position $t$}
            \STATE Set context
            $s_{i,t}=(x_i,y_{i,<t})$
            \STATE Obtain teacher Top-$K$ set
            $\mathcal{K}_{i,t}$
            \STATE Store $\log\pi_T(v\mid s_{i,t})$
            for $v\in\mathcal{K}_{i,t}$
        \ENDFOR
    \ENDFOR

    \STATE \textbf{// Student distribution matching}
    \STATE Forward $\pi_\theta$ on
    $\{z_i\}_{i=1}^{B}$

    \FOR{response positions in chunks of size $C$}
        \STATE Compute student log-normalizers
        \STATE Gather logits at teacher Top-$K$ indices
        \STATE Obtain $\log\pi_\theta(v\mid s_{i,t})$
        for $v\in\mathcal{K}_{i,t}$
    \ENDFOR

    \STATE Clip teacher and student log-probabilities
    from below at $-10$

    \STATE \textbf{// Top-$K$ forward-KL distillation}
    \FOR{each valid response position $(i,t)$}
        \STATE Compute $\ell_{i,t}$ using
        Eq.~\eqref{eq:opd_topk_kl}
        \STATE Clip $\ell_{i,t}$ to $[0,10]$
    \ENDFOR

    \STATE Aggregate token losses using
    Eq.~\eqref{eq:opd_loss}

    \STATE Set
    $\mathcal{L}_{\mathrm{total}}
    =\mathcal{L}_{\mathrm{OPD}}$

    \STATE Update
    $\theta \leftarrow
    \theta-\eta\nabla_\theta
    \mathcal{L}_{\mathrm{total}}$

    \STATE Synchronize student weights
    to the rollout engine

\ENDFOR

\end{algorithmic}
\end{algorithm}

Unlike policy-gradient OPD, our configuration directly backpropagates the
distribution-matching objective through the student probabilities. Task rewards
and PPO/GRPO policy-gradient losses are disabled; hence the total optimization
objective is simply
\begin{equation}
    \mathcal{L}_{\mathrm{total}}
    =
    \mathcal{L}_{\mathrm{OPD}}.
\end{equation}
Because each new response is sampled from the latest student before teacher
scoring, the distillation states evolve together with the student policy,
rather than remaining fixed to an offline teacher-generated dataset.

\section{Implementation Details}
We first train Qwen3-VL-8B-Instruct with curriculum SFT using supervision
constructed by Qwen3.7-Max. The resulting SFT checkpoint initializes
\textsc{Skill OPD}, in which a fixed Qwen3-VL-32B-Instruct teacher
provides token-level soft targets for the editing rollouts generated by
the current student policy. We use pure Top-$K$ forward-KL distillation
without auxiliary reward or policy-gradient objectives. The complete
data-generation, optimization, batching, sequence-length, and system
configurations are reported in Table~\ref{tab:training_hyperparameters}. The training time for SFT stages are about 7 GPU hours in total and 64 GPU hours for OPD. So overall, the method is computationally lightweight. 

\begin{table}[!t]
\centering

\setlength{\tabcolsep}{4.0pt}
\renewcommand{\arraystretch}{1.08}
\small
\begin{tabular}{
    @{}p{0.47\columnwidth}
    p{0.47\columnwidth}@{}
}
\toprule
\textbf{Hyperparameter} & \textbf{Value} \\
\midrule

\multicolumn{2}{c}{\textit{Curriculum SFT}} \\
\midrule

Student model
& Qwen3-VL-8B-Instruct \\

Supervision model
& Qwen3.7-Max \\

Training epochs
& 3 \\

Optimizer
& AdamW \\

Learning rate
& $5\times10^{-7}$ \\

Learning-rate schedule
& Cosine \\

Weight decay
& $0.01$ \\

Gradient clipping
& $1.0$ \\

Maximum sequence length
& $8{,}192$ \\

Training precision
& \texttt{bfloat16} \\

Stage~0 Training time
& $\sim$ 4 GPU hours \\

Stage~1 Training time
& $\sim$ 3 GPU hours \\

\midrule
\multicolumn{2}{c}{\textit{\textit{Skill OPD}}} \\
\midrule

Student model
& Stage 1 SFT checkpoint \\

Teacher model
& Qwen3-VL-32B-Instruct \\

Teacher update
& Frozen \\

Distillation objective
& Top-$K$ forward KL \\

Teacher Top-$K$
& 32 \\

Top-$K$ chunk size
& 8 \\

Distillation loss coefficient
& $1.0$ \\

Maximum loss clipping
& $10$ \\

Minimum log-prob. clipping
& $-10$ \\

Training epochs
& 6 \\

Training time
& $\sim$ 64 GPU hours \\

Global rollout batch size
& 8 \\

Rollouts per prompt $N$
& 1 \\

Optimization mini-batch size
& 8 \\

Optimization epochs per rollout batch
& 1 \\

Policy updates per rollout batch
& 1 \\

Per-GPU micro-batch size
& 1 \\

Optimizer
& AdamW \\

Learning rate
& $5\times10^{-7}$ \\

Learning-rate schedule
& Constant \\

Weight decay
& $0.01$ \\

Gradient clipping
& $1.0$ \\

Per-GPU token budget
& $12{,}289$ \\

Student training resources
& FSDP on $8\times$ A100-40GB\\

Student rollout
& vLLM Colocated, TP${}=2$ \\

Teacher serving resources
& vLLM on $8\times$ A100-40GB, TP${}=8$\\

Rollout precision
& \texttt{bfloat16} \\

vLLM memory utilization
& $0.6$ \\

\bottomrule
\end{tabular}
\caption{
Training hyperparameters and system configurations for curriculum SFT
and \textit{Skill OPD}.
}
\label{tab:training_hyperparameters}
\end{table}

\section{Skill Analysis}
\label{sec:skill_execution_analysis}

\subsection{What We Learned}

\begin{figure}[!t]
    \centering
    \includegraphics[width=0.8\linewidth]{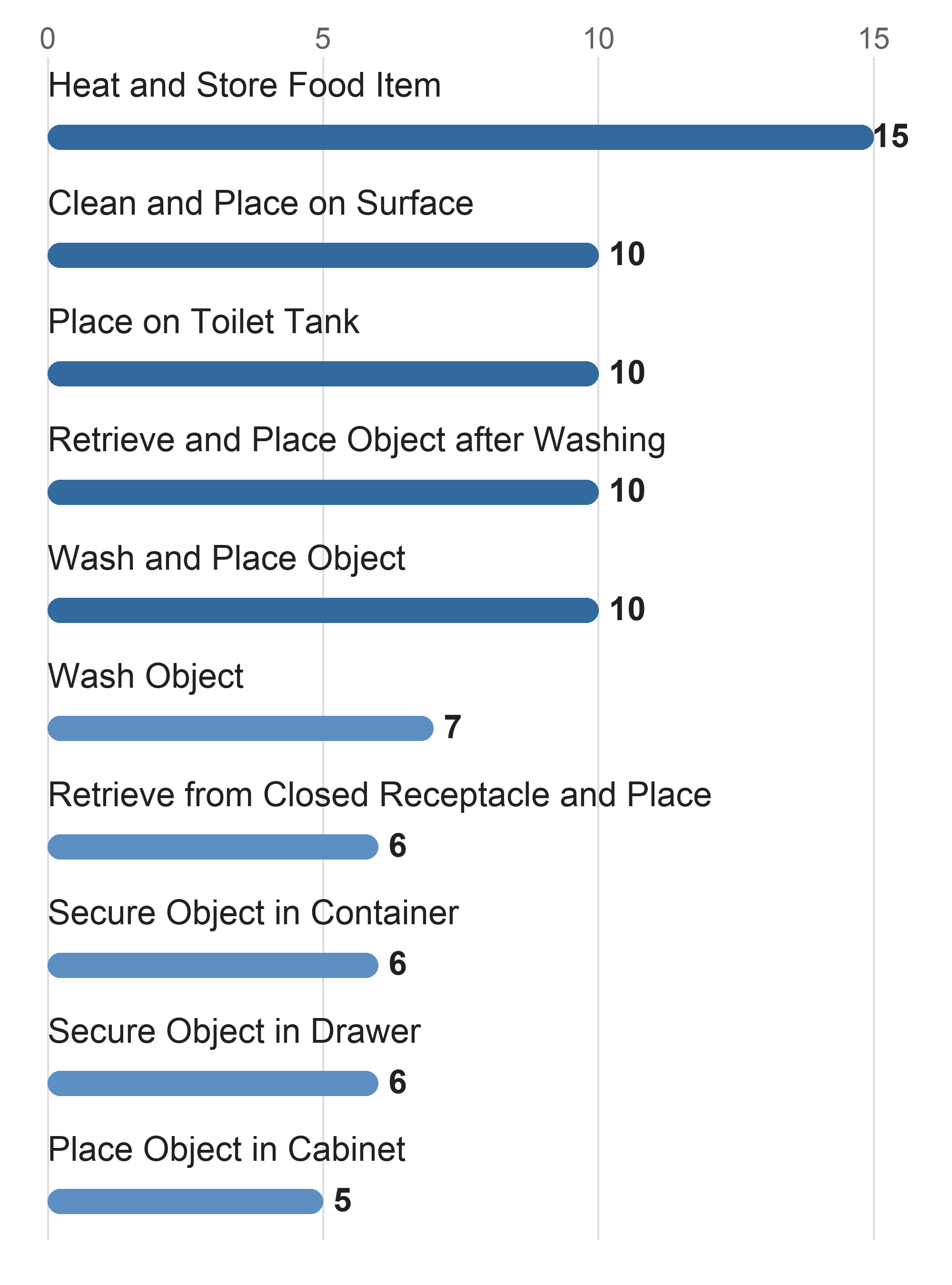}
    \caption{The 10 skills with the longest default action sequences in the EB-ALFRED skill library.
}
    \label{fig:skill_sequence_lengths_alfred}
\end{figure}

\begin{figure}[!t]
    \centering
    \includegraphics[width=0.8\linewidth]{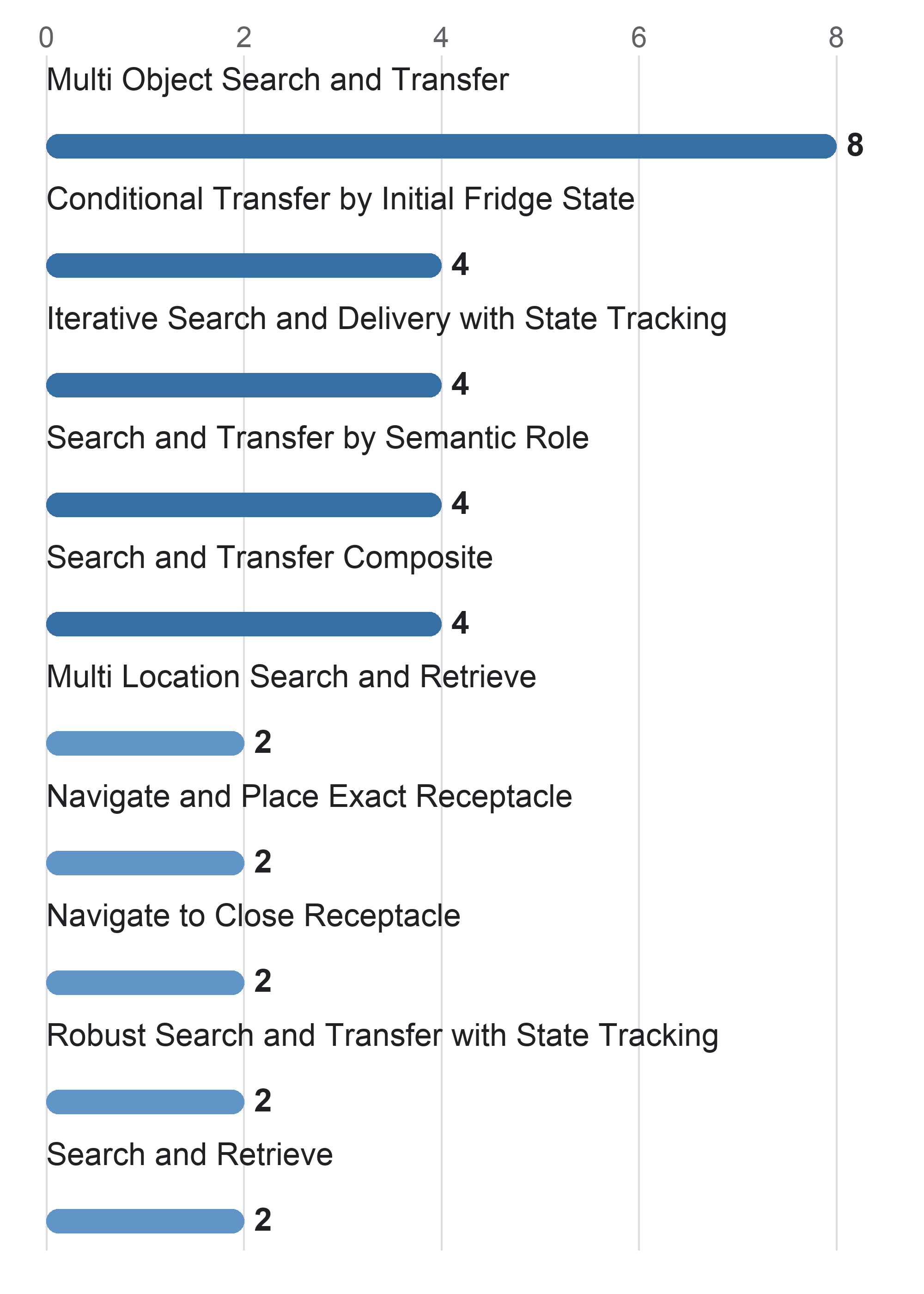}
    \caption{The 10 skills with the longest default action sequences in the EB-Habitat skill library.
}
    \label{fig:skill_sequence_lengths_habitat}
\end{figure}

We analyze the action-sequence lengths of the skill libraries produced by the skill learner after all three training stages on both EB-ALFRED and EB-Habitat. As shown in Figure~\ref{fig:skill_sequence_lengths_alfred}-\ref{fig:skill_sequence_lengths_habitat}, we visualize the ten longest skills for each benchmark. On EB-ALFRED, the learned library contains 29 skills, comprising 13 single-action skills and 16 composite skills. On EB-Habitat, it contains 15 skills, including 5 single-action skills and 10 composite skills.

The learner discovers reusable procedures at substantially different levels of complexity. On EB-ALFRED, \texttt{HeatAndStoreFoodItem} expands into 15 actions, integrating object retrieval, appliance operation, heating, and final storage. On EB-Habitat, \texttt{MultiObjectSearchAndTransfer} contains 8 actions and captures repeated object search, retrieval, and delivery. These examples demonstrate that \projname{} can abstract complete long-horizon procedures. Such long macros can simplify planning, especially those compex long-horizon tasks.

\begin{figure}[!t]
    \centering
    \includegraphics[width=0.9\linewidth]{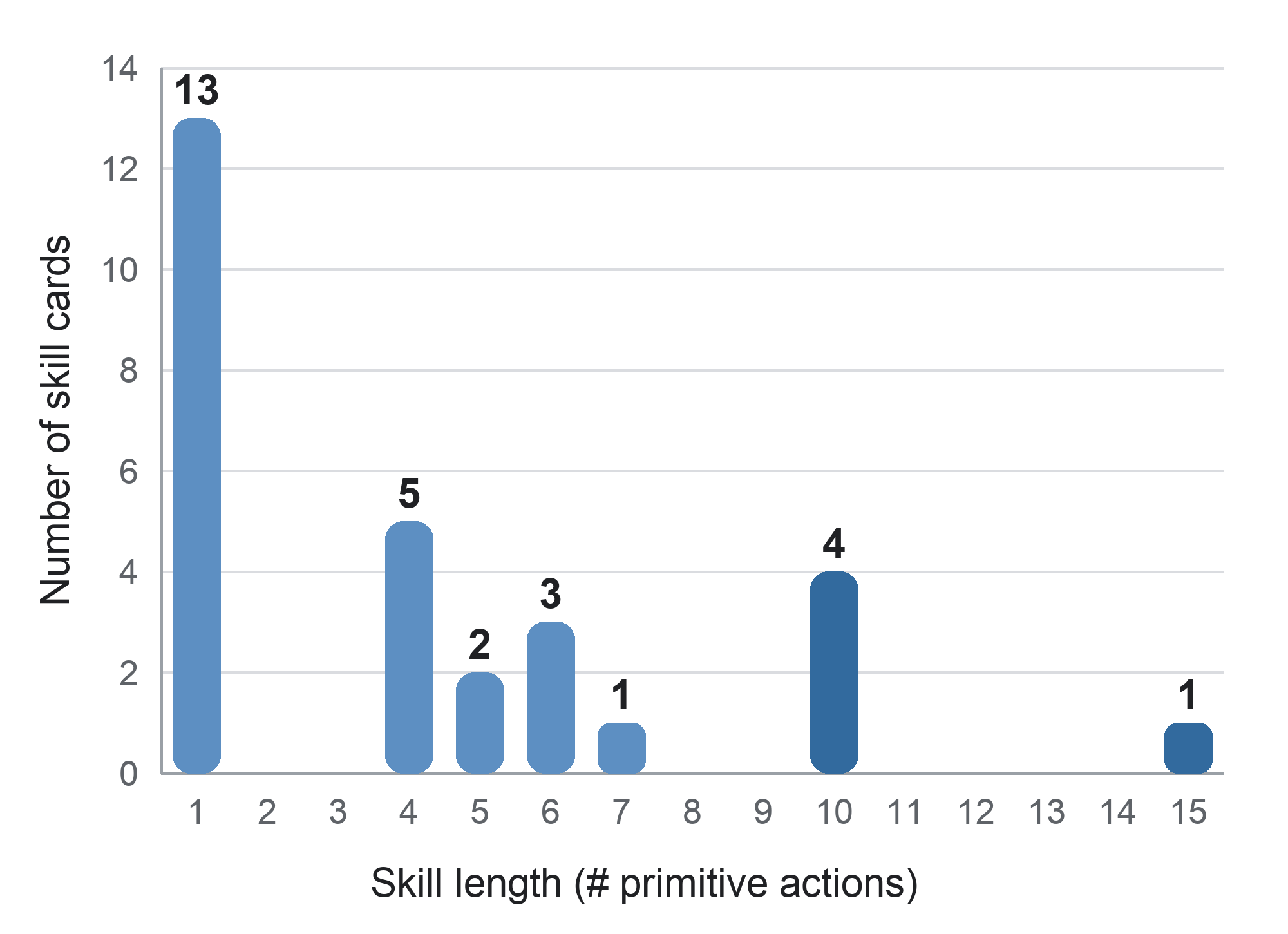}
    \includegraphics[width=0.9\linewidth]{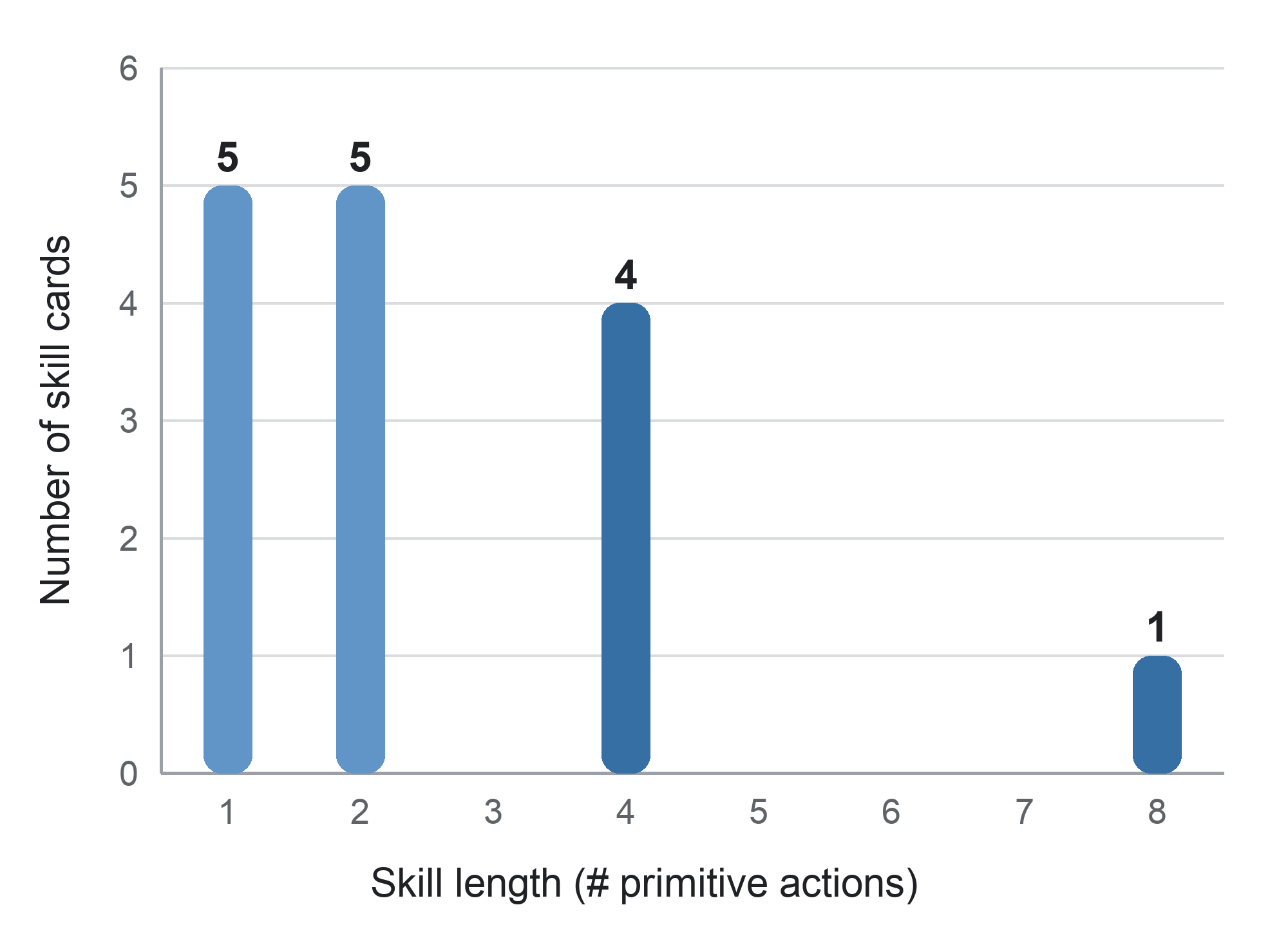}
    \caption{
Distribution of default skill-expansion lengths in the learned
skill libraries for (a) EB-ALFRED and (b) EB-Habitat. Skill length
is measured by the number of primitive actions in the default
expansion pattern.
}
    \label{fig:skill_sequence_lengths_distribution}
\end{figure}

As shown in the Figure~\ref{fig:skill_sequence_lengths_distribution}, both libraries exhibit a clear hierarchy between primitive control and procedural abstraction. In EB-ALFRED, 11 of the 16 composite skills contain 4 to 7 actions. In EB-Habitat, 9 of the 10 composite skills contain 2 to 4 actions, with only one extending to 8 actions. This difference indicates that \projname{} adapts skill granularity to the procedural structure of each environment: EB-ALFRED requires longer manipulation routines, whereas EB-Habitat primarily benefits from compact search-and-transfer abstractions.


\subsection{Skill Execution Analysis}

\begin{figure}[!t]
    \centering
    \includegraphics[width=0.7\linewidth]{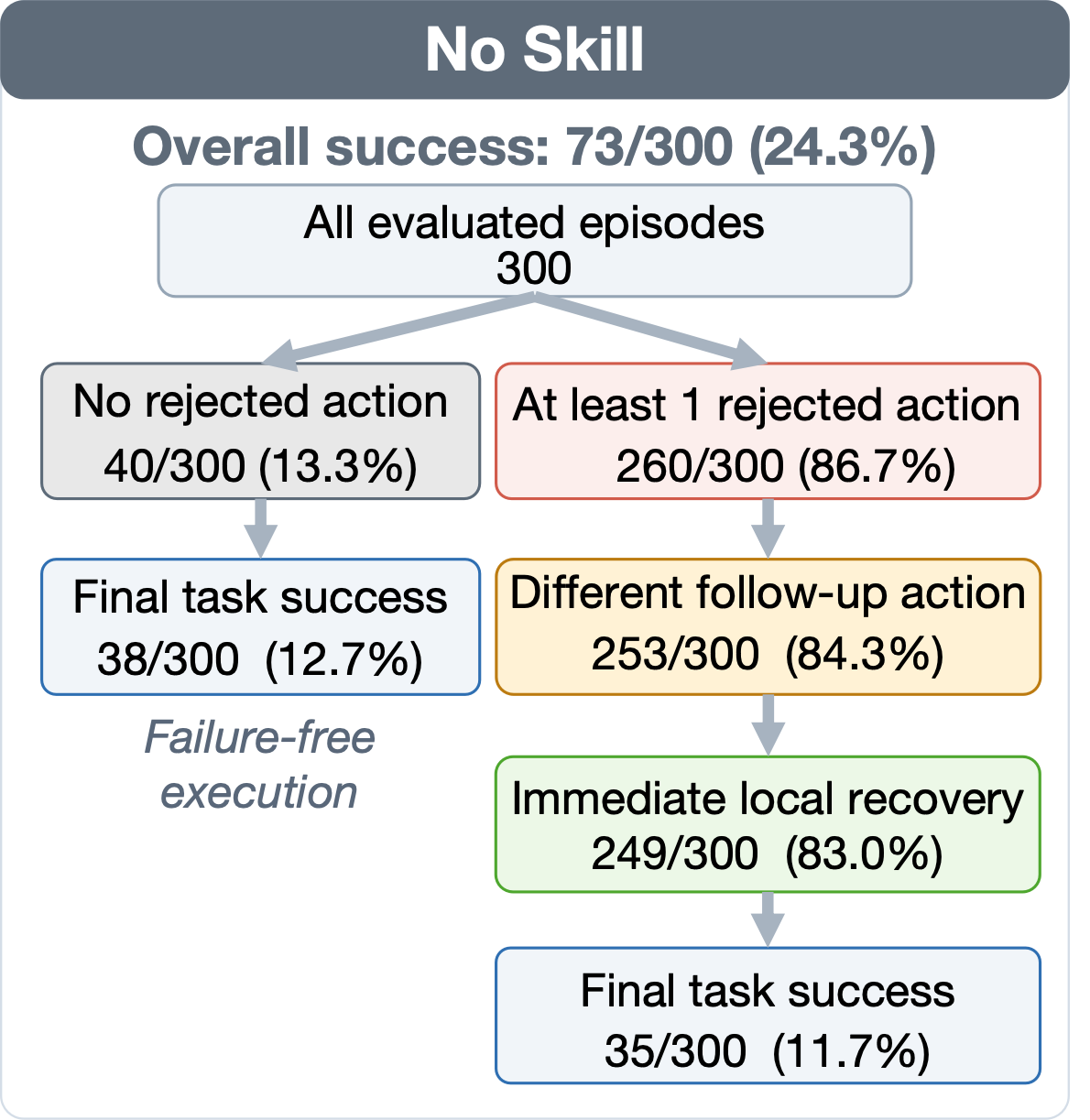}
    \includegraphics[width=0.7\linewidth]{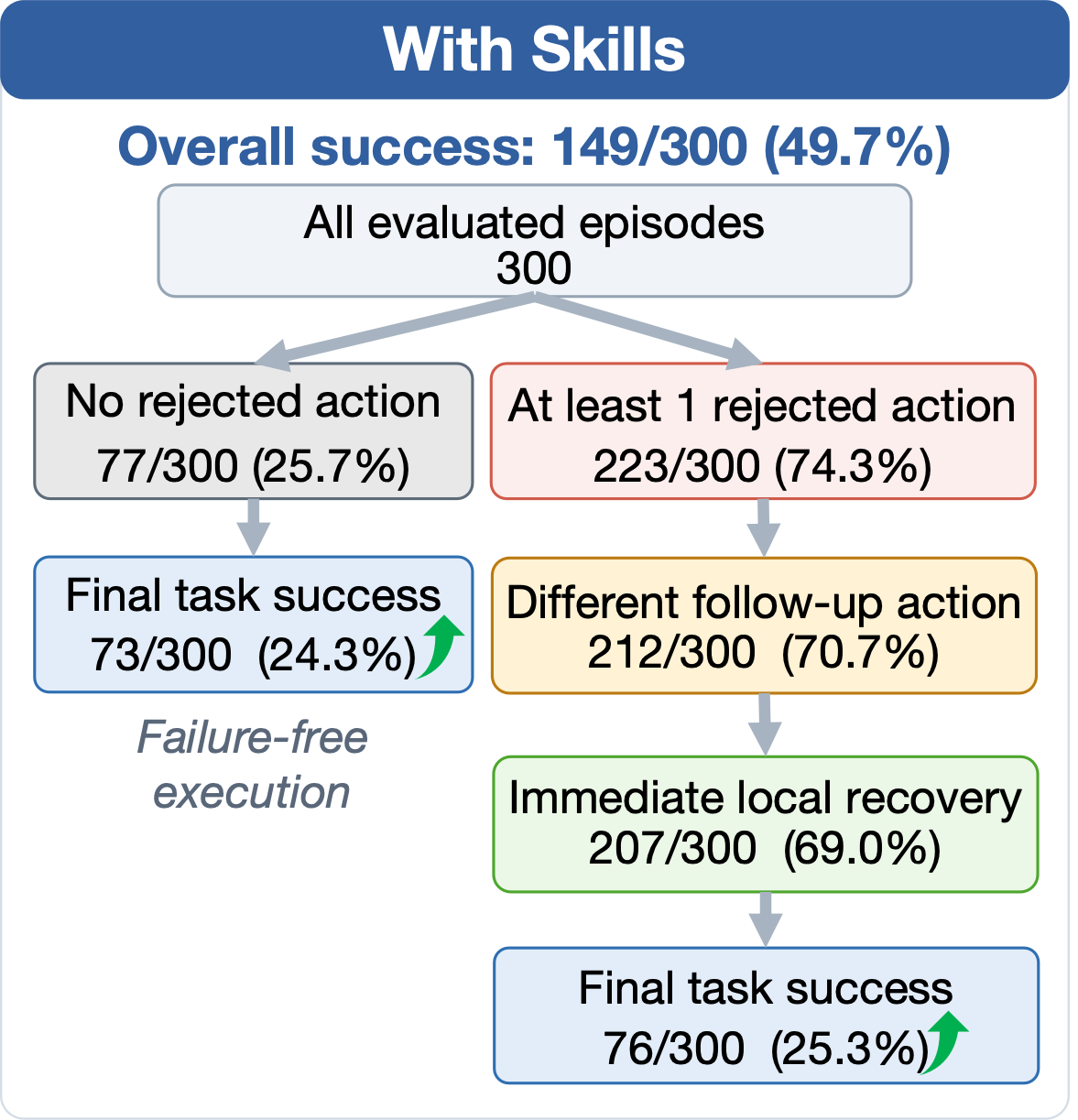}
    \caption{
    Execution outcomes with and without learned skills over 300 EB-ALFRED episodes.
    }
    \label{fig:full300_recovery_funnel}
\end{figure}

We analyze how the same executor behaves with and without the learned skill library. Figure~\ref{fig:full300_recovery_funnel} summarizes all 300 EB-ALFRED episodes. First, learned skills enable more failure-free execution. The number of episodes without any rejected action increases from 40/300 (13.3\%) to 77/300 (25.7\%), while successful failure-free episodes increase from 38/300 (12.7\%) to 73/300 (24.3\%). This indicates that the learned skills help the executor generate more executable plans that satisfy action preconditions.

Second, learned skills substantially improve outcomes even when rejected actions still occur. Although the number of episodes containing a rejected action decreases from 260 to 223, successful episodes within this branch increase from 35/300 (11.7\%) to 76/300 (25.3\%). The lower population-level percentages for different follow-up actions and immediate local recovery mainly reflect that fewer episodes enter the failure branch; they should not be interpreted as conditional recovery rates.

Overall, the improvement is jointly explained by proactive failure avoidance and post-failure recovery. These results show that \projname{} improves embodied task execution not only by producing more valid plans, but also by helping the executor preserve progress and ultimately complete the task after execution failures.

\begin{figure*}[!t]
    \centering
    \includegraphics[width=0.9\linewidth]{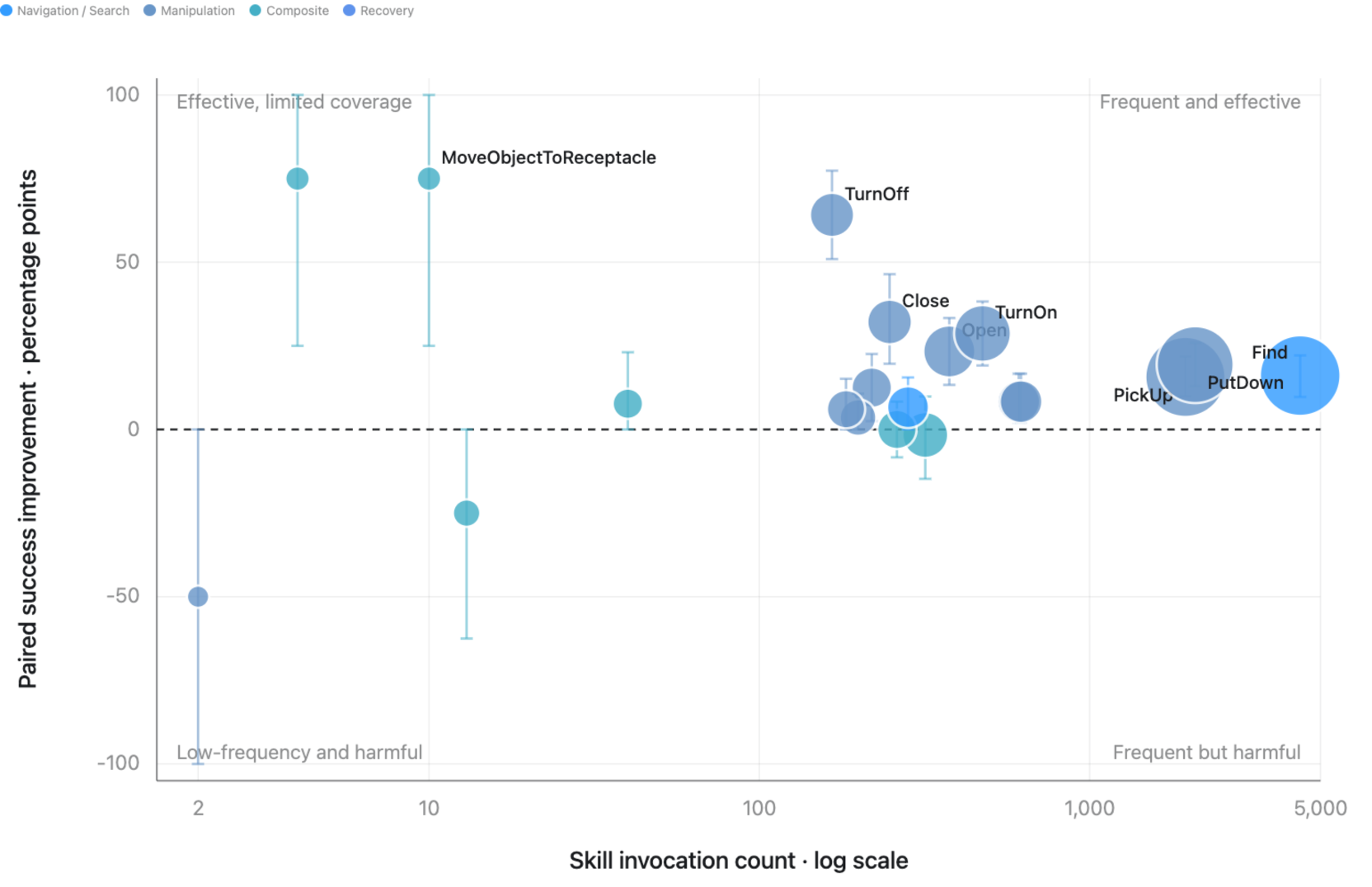}
    \caption{Skill-level benefit–risk analysis. Invocation frequency is plotted against paired success-rate improvement over the no-skill baseline; bubble size indicates task coverage, color denotes skill family, and error bars show 95\% bootstrap confidence intervals.}
    \label{fig:skill_utilization}
\end{figure*}

We further analyze the execution situation of each skills. The Figure~\ref{fig:skill_utilization} illustrates the relationship between skill usage frequency and task-level benefit. The x-axis shows the number of skill invocations, while the y-axis reports the paired success-rate improvement over the no-skill 0-shot baseline on the same episodes. Bubble size represents task coverage, and error bars indicate 95\% bootstrap confidence intervals. Compared with success rates conditioned only on skill invocation, this paired comparison better controls for episode difficulty and reduces selection bias.

Overall, the \projname{} method improves the task success rate from 24.3\% to 40.0\%. \texttt{TurnOff}, \texttt{Close}, \texttt{TurnOn}, \texttt{Open}, \texttt{PutDown}, \texttt{Find}, and \texttt{PickUp} all show consistent positive improvements. In particular, \texttt{Find}, \texttt{PickUp}, and \texttt{PutDown} combine high invocation frequency with broad task coverage, demonstrating strong generalizability. \texttt{TurnOff}, \texttt{Close}, and \texttt{TurnOn} achieve larger gains, suggesting that skill cards effectively stabilize interaction procedures that are otherwise error-prone. \texttt{MoveObjectToReceptacle} and \texttt{WashObject} also show large improvements, although their limited coverage requires further validation.

The four quadrants represent distinct types of skills. The upper-right quadrant contains frequent, broadly applicable, and effective skills, which form the core strength of the learned skill library. The upper-left quadrant contains promising specialized skills with insufficient coverage and therefore calls for additional evaluation data. The lower-right quadrant represents frequently used but harmful skills; this region is nearly empty, indicating that high-frequency skills do not introduce systematic negative effects. The lower-left quadrant mainly contains low-frequency skills such as \texttt{Drop} and \texttt{HoldObjectWhileTurningOnLight}, whose negative estimates remain uncertain because of limited samples. Skills such as \texttt{PlaceObjectInContainer}, \texttt{PlaceObjectOnNonContainerSurface}, \texttt{TurnOffLight}, and \texttt{SliceWithTool} do not show robust improvements and should be prioritized for expansion or routing refinement.

Overall, the learned skill library improves success while concentrating model usage on broadly reusable skills with stable positive returns. These results demonstrate that the proposed method can extract transferable behavioral patterns from trajectories and improve execution reliability across diverse tasks.

\begin{figure}[!t]
    \centering
    \includegraphics[width=\linewidth]{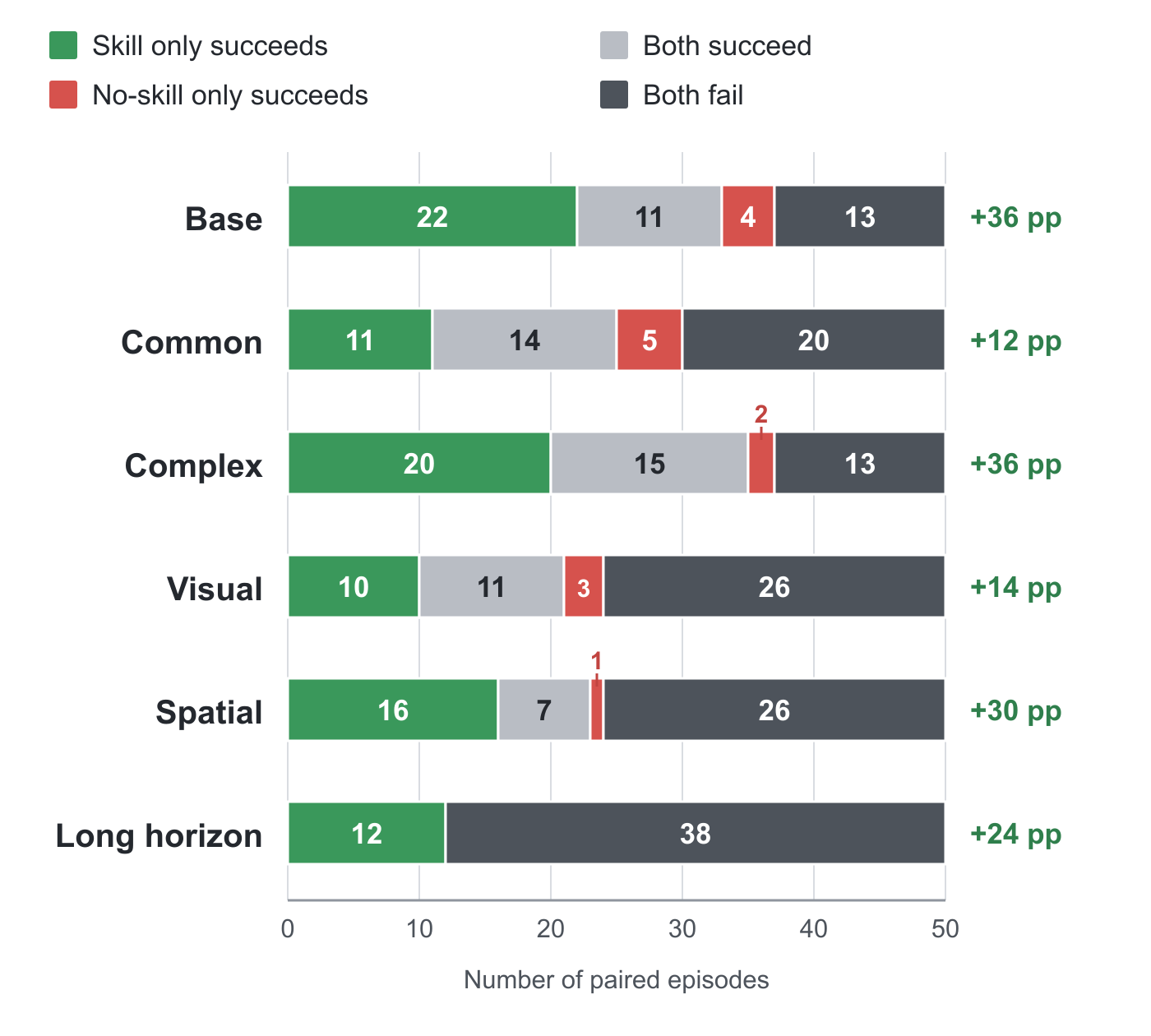}
    \caption{Paired episode outcomes between the no-skill baseline and \projname{} across six EB-ALFRED splits. Each bar shows the four paired outcome categories, with the corresponding change in success rate (\(\Delta\mathrm{SR}\)) reported on the right.}
    \label{fig:paired_skill_outcomes}
\end{figure}

\begin{figure}[!t]
    \centering
    \includegraphics[width=\linewidth]{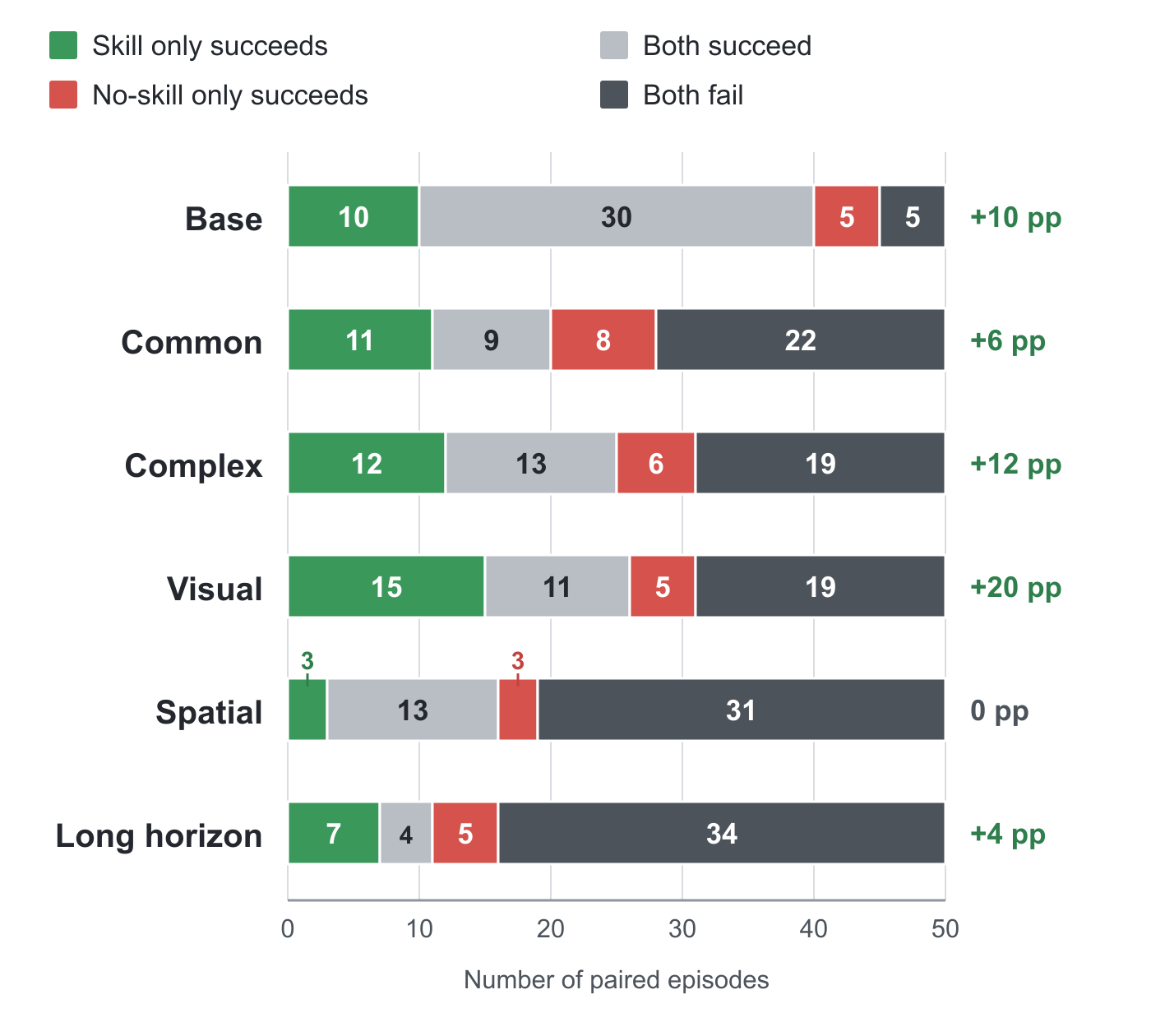}
    \caption{Paired episode outcomes between the no-skill baseline and \projname{} across six EB-Habitat splits. Each bar shows the four paired outcome categories, with the corresponding change in success rate (\(\Delta\mathrm{SR}\)) reported on the right.}
    \label{fig:paired_skill_outcomes_2}
\end{figure}

\subsection{Skill Usage in Different Splits}

We compare the \projname{} with the No-Skill baseline through paired evaluation through six different splits. Each episode is categorized as both agents succeeding, only the \projname{} succeeding, only the No-Skill agent succeeding, or both agents failing.

As shown in Figure~\ref{fig:paired_skill_outcomes}-\ref{fig:paired_skill_outcomes_2}, the \projname{} provides consistent gains on both benchmarks. On EB-ALFRED, skills recover 91 episodes that the No-Skill baseline fails, while introducing regressions on only 15 episodes, resulting in a net improvement of $\Delta\mathrm{SR}=+25.3$ percentage points. The largest gains appear on base and complex-instruction tasks ($+36$ points), followed by spatial tasks ($+30$ points). On long-horizon tasks, the \projname{} solves 12 episodes while the baseline solves none. These results indicate that reusable skills effectively translate diverse instructions into stable multi-step action sequences.

On EB-Habitat, skills recover 58 failed episodes while causing 32 regressions, producing an overall improvement of $\Delta\mathrm{SR}=+8.7$ points. The largest gain occurs on visual-appearance tasks ($+20$ points), followed by complex-instruction ($+12$ points) and base tasks ($+10$ points). The smaller improvement on EB-Habitat reflects its stronger dependence on scene exploration and object localization: navigation is restricted to receptacles, so successful execution often requires locating an object indirectly before applying a reusable manipulation pattern. Spatial tasks show no net improvement, while long-horizon tasks improve by only 4 points, suggesting that spatial grounding and extended state tracking remain difficult to encode through fixed skill structures alone.

Overall, skills provide a substantial paired advantage on both benchmarks. Their primary benefit is reducing the planning complexity of recurring multi-step behaviors and improving execution consistency. Remaining failures are concentrated in tasks requiring visual grounding, indirect object search, spatial reasoning, and long-horizon coordination, highlighting the need for stronger adaptive skill selection and state-aware composition.

\begin{table}[!t]
\centering
\caption{Skill execution and recovery statistics on EB-ALFRED. 
Inv.: skill invoked; Exp.: valid expansion; Acc.: primitive-action acceptance; 
Fail: episodes with rejected actions; Rec.: locally recovered episodes; 
Succ.: task success. Each split contains 50 episodes.}
\label{tab:skill_execution_by_split}
\setlength{\tabcolsep}{2.5pt}
\renewcommand{\arraystretch}{1.05}
\resizebox{\columnwidth}{!}{
\begin{tabular}{lcccccc}
\toprule
\textbf{Split} & \textbf{Inv.} & \textbf{Exp.} & \textbf{Acc.}
& \textbf{Fail} & \textbf{Rec.} & \textbf{Succ.} \\
\midrule
Base & 50/50 & 47/50 & 82.6\% & 31/50 & 30/50 & 66.0\% \\
CS   & 49/50 & 42/50 & 81.2\% & 36/50 & 32/50 & 50.0\% \\
CI   & 49/50 & 48/50 & 85.8\% & 31/50 & 29/50 & 70.0\% \\
VA   & 49/50 & 41/50 & 74.6\% & 40/50 & 34/50 & 42.0\% \\
SR   & 50/50 & 44/50 & 76.5\% & 40/50 & 37/50 & 46.0\% \\
LH   & 50/50 & 48/50 & 80.5\% & 45/50 & 45/50 & 24.0\% \\
\midrule
All  & 297/300 & 270/300 & 80.2\% & 223/300 & 207/300 & 49.7\% \\
\bottomrule
\end{tabular}
}
\end{table}
Table~\ref{tab:skill_execution_by_split} analyzes skill invocation, action expansion, execution validity, and failure recovery across different splits on EB-ALFRED. The best results occur on Complex Instruction and Base, which achieve the highest task success rates (70.0\% and 66.0\%) together with reliable skill expansion and action acceptance. This suggests that reusable skills effectively convert diverse language instructions into stable action structures. Common Sense remains moderately successful, while Visual Appearance and Spatial Relationship are more difficult because correct execution depends heavily on visual instance identification and spatial grounding. Long Horizon exhibits high expansion validity and recovers from every local failure, yet reaches only 24.0\% task success. This indicates that local recovery is effective, but cannot fully prevent error accumulation or maintain state consistency over extended skill compositions. Overall, the results highlight the method's strong execution and recovery capabilities while identifying visual grounding and long-horizon coordination as the main remaining bottlenecks.

\FloatBarrier
\section{Qualitative Analysis}
\label{sec:qualitative_analysis}

We qualitatively examine representative successful and failed trajectories to understand how the learned skill library supports execution and where its limitations arise. We focus on whether reusable skills reduce planning complexity, how the executor responds to rejected actions, and whether failures originate from the skill abstraction or from online grounding and execution.

\subsection{Case Study in EB-ALFRED}
\label{sec:case_study_alfred}

\begin{figure*}[!t]
    \centering
    \includegraphics[width=0.49\textwidth]{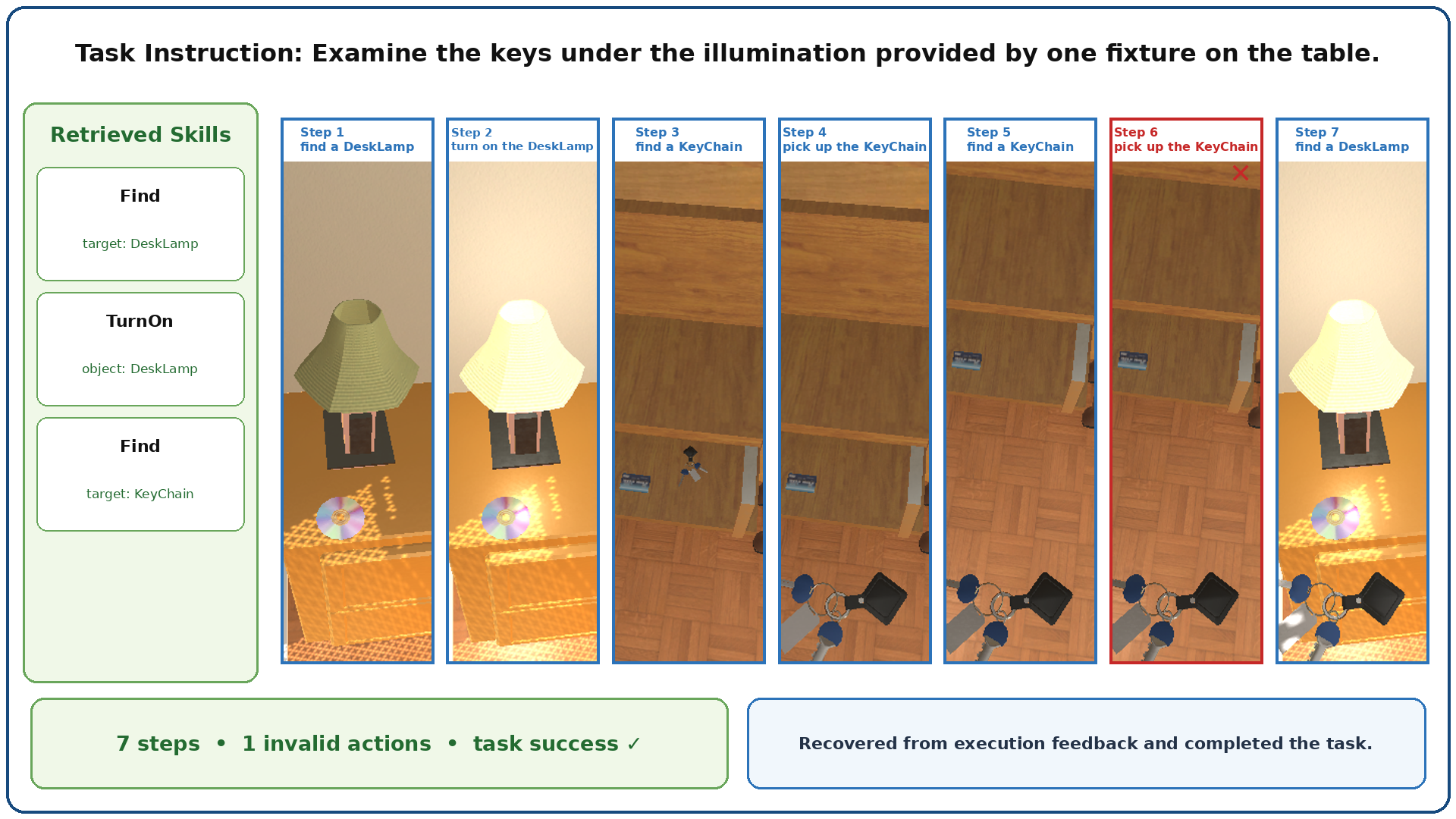}
    \hfill
    \includegraphics[width=0.49\textwidth]{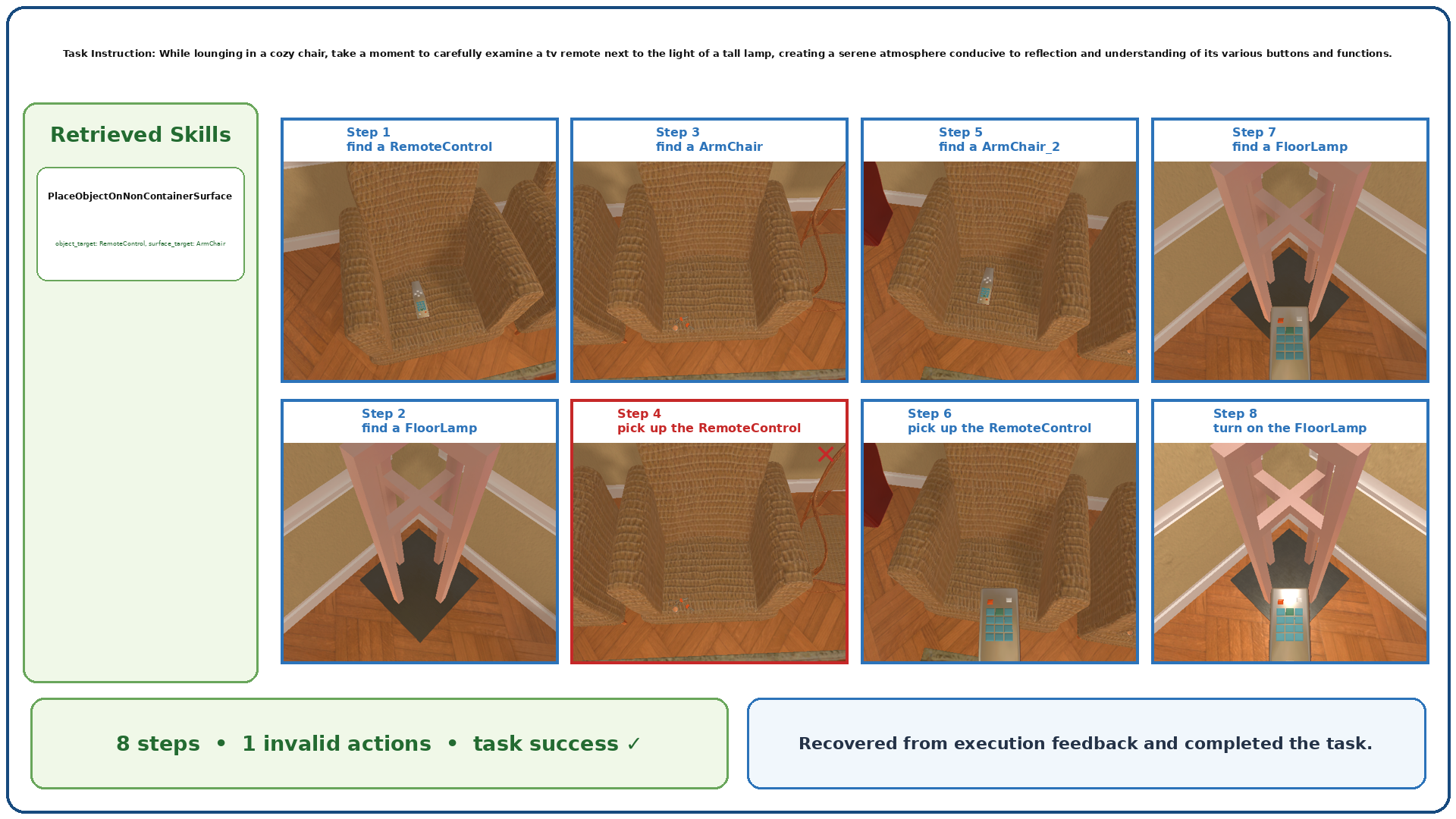}
    \caption{Successful light-assisted inspection in EB-ALFRED. Common Sense-49 (left) establishes illumination and recovers from a rejected keychain pickup, while Complex-30 (right) combines reusable object placement with lamp activation.}
    \label{fig:alfred_light_inspection}
\end{figure*}

\paragraph{Successful cases: light-assisted inspection.}
Light-assisted inspection requires the agent to coordinate object localization, illumination, and manipulation. Figure~\ref{fig:alfred_light_inspection} shows two representative examples. In Common Sense-49, the agent grounds the indirect expressions ``keys'' and ``one fixture on the table'' to \texttt{KeyChain} and \texttt{DeskLamp}. It first invokes \textsc{TurnOn} to establish the required lighting condition and then uses \textsc{Find} and \textsc{PickUp} to inspect the keys. After one rejected pickup, the executor re-localizes the keychain and completes the task without restarting the plan. In Complex-30, the composite skill \textsc{PlaceObjectOnNonContainerSurface} grounds the remote control and armchair as a reusable placement subtask, while the executor additionally activates the floor lamp. Together, these examples demonstrate that the method can combine explicit composite skills with feedback-aware primitive execution to efficiently solve semantically diverse inspection tasks.

\begin{figure*}[!t]
\centering
\includegraphics[width=0.49\textwidth]{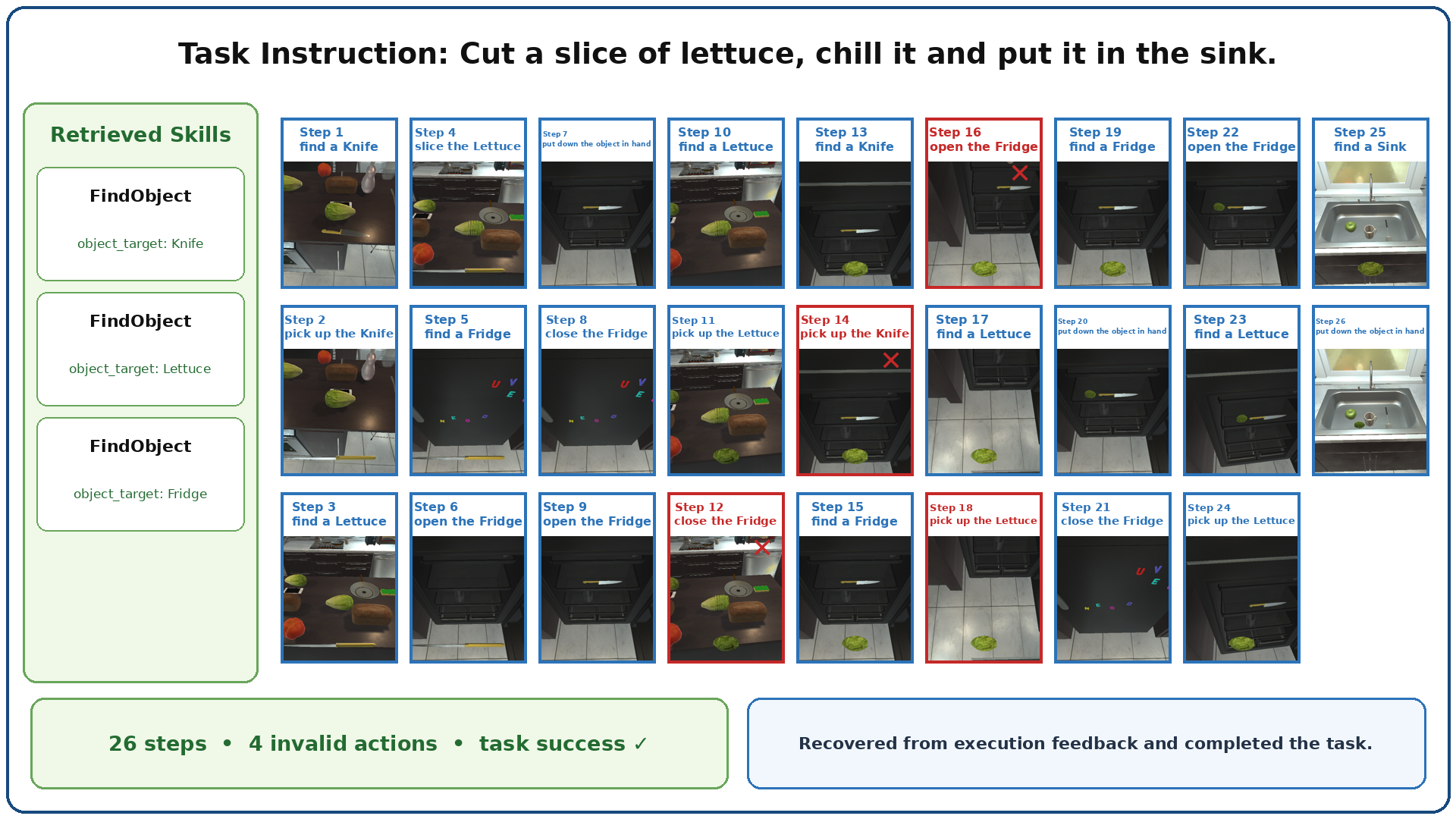}
\hfill
\includegraphics[width=0.49\textwidth]{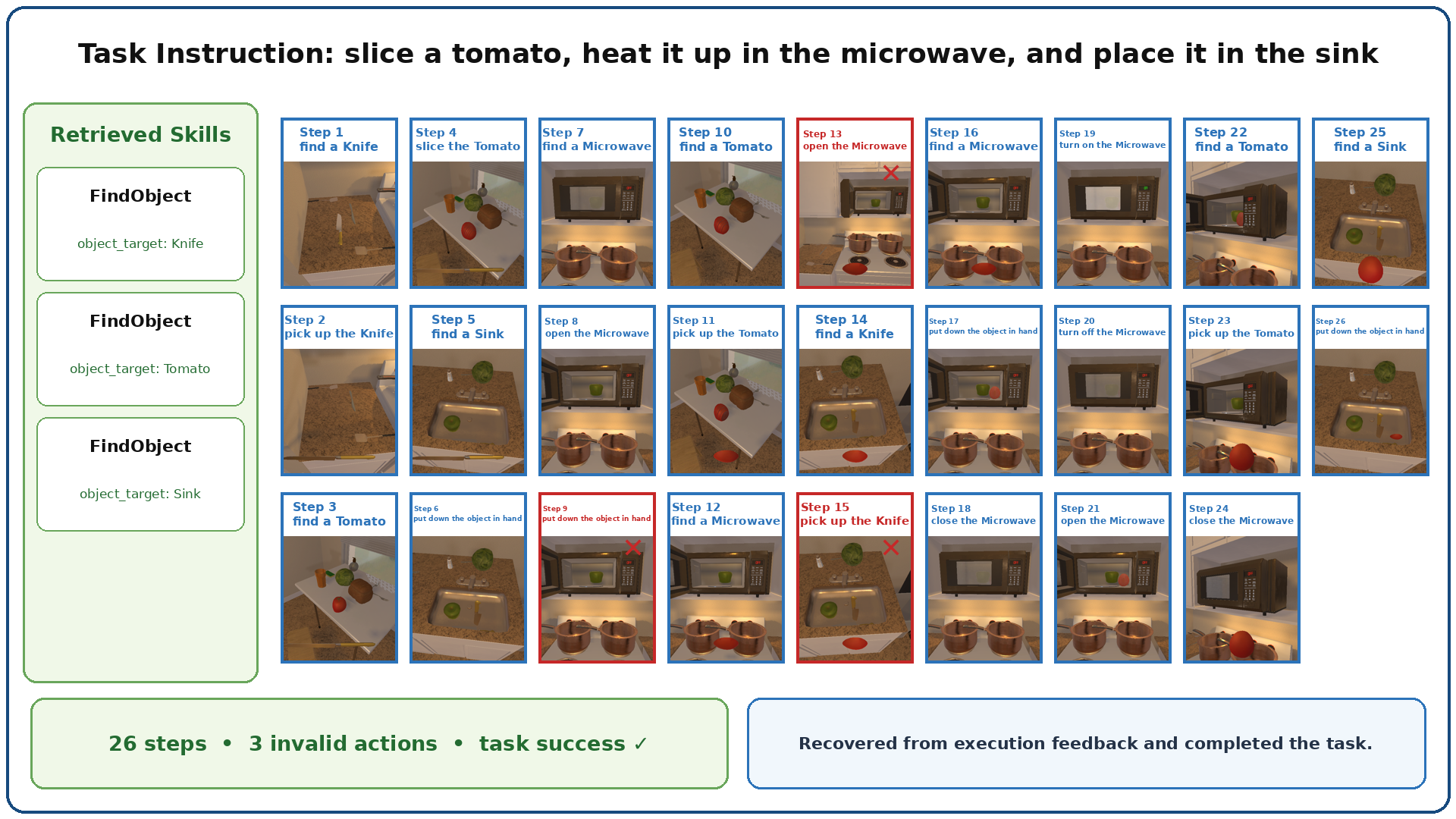}
\caption{Successful long-horizon transformations in EB-ALFRED. The agent composes reusable object-location skills with slicing, cooling or heating, and final placement.}
\label{fig:alfred_long_horizon_success}
\end{figure*}

\paragraph{Successful cases: observable long-horizon transformations.}Figure~\ref{fig:alfred_long_horizon_success} presents two long-horizon tasks involving object transformation and relocation. Long-27 retrieves reusable \textsc{FindObject} cards for the knife, lettuce, and refrigerator, and composes them with slicing, cooling, and final placement actions. Long-45 similarly uses object-location skills as anchors for slicing a tomato, heating it in the microwave, and placing it in the sink. These trajectories do not rely on a single monolithic task program. Instead, the executor composes reusable skills with state-changing primitive actions and continues after several rejected actions. This modular structure allows it to preserve completed subgoals and recover locally rather than restarting the entire sequence.

\paragraph{Failure cases: containment and holding state.}
Tasks requiring one object to be placed inside another are particularly sensitive to holding-state consistency. In Base-07, the agent must insert a butter knife into a cup and then move the cup to the sink. It alternates between the two objects, repeatedly attempts pickup while holding another object, and loses the required containment relation. Complex-35 retrieves \textsc{PlaceObjectOnNonContainerSurface} for placing an apple in a pan. This skill is insufficient because the task additionally requires transporting the filled pan into the refrigerator. The executor consequently alternates between the apple and pan without establishing a stable nested state. These cases expose both incomplete skill coverage for nested transport and insufficient executor-side tracking of the object in hand.

\begin{figure*}[!t]
    \centering
    \includegraphics[width=0.49\textwidth]{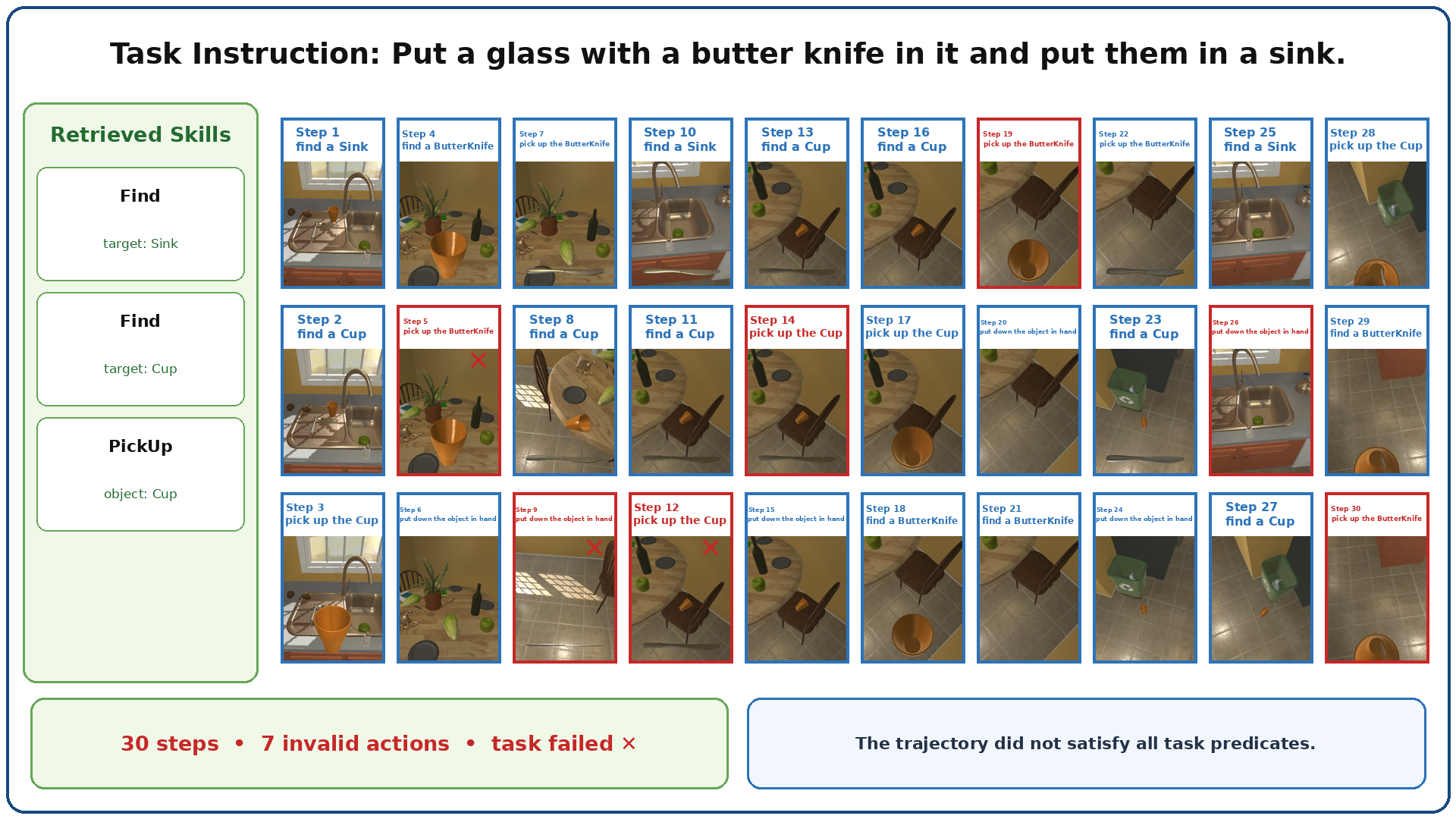}
    \hfill
    \includegraphics[width=0.49\textwidth]{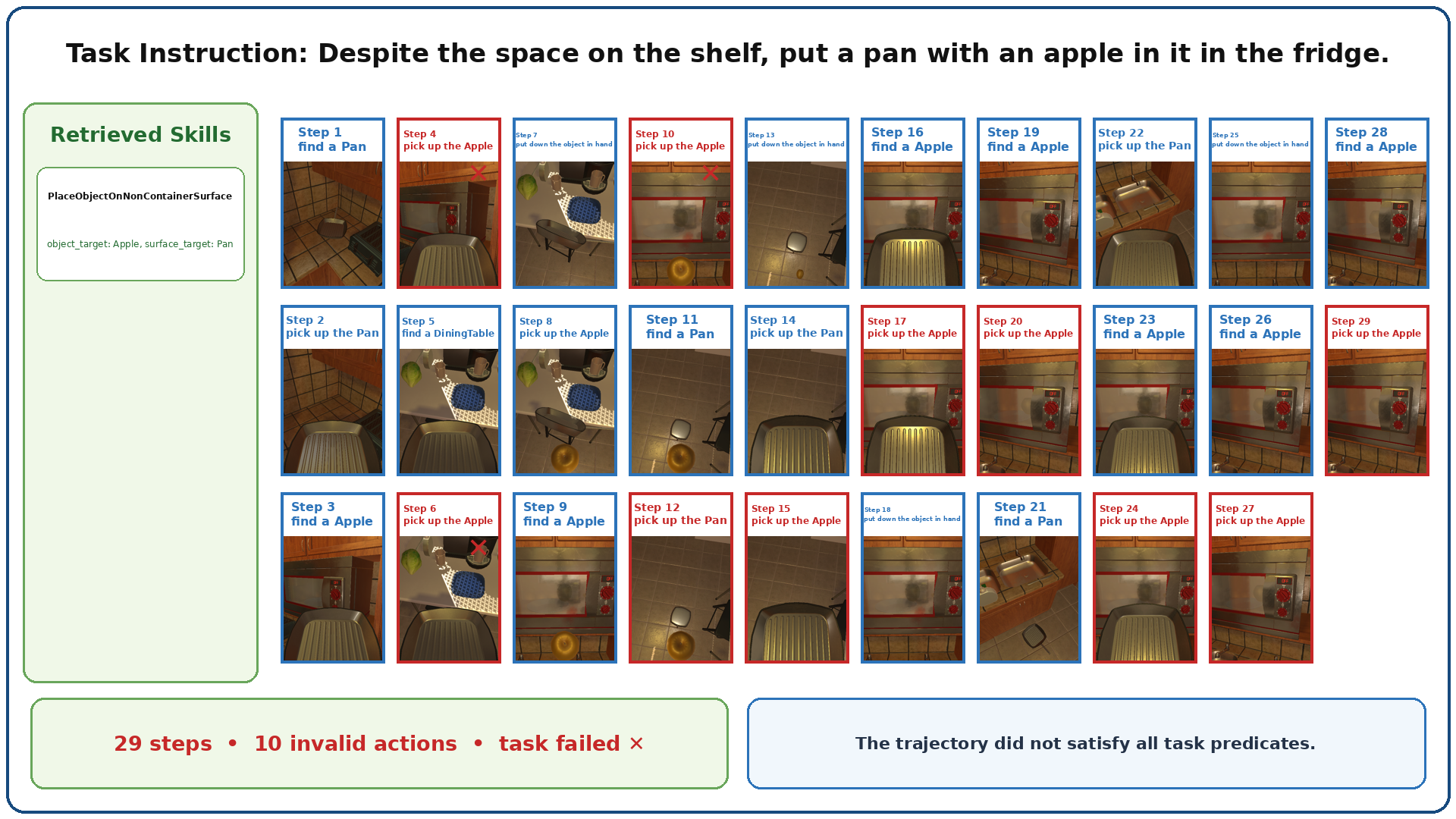}
    \caption{Containment and holding-state failures in EB-ALFRED. The executor fails to preserve the required cup--knife (left) and pan--apple (right) relations.}
    \label{fig:alfred_holding_failures}
\end{figure*}

Overall, the learned skills compress recurring multi-step behaviors into stable execution structures and enable recovery from isolated local errors. The remaining limitations largely lie outside basic skill expansion: dynamic grounding, persistent object identity, holding-state management, and long-term consistency across multiple skill calls.

\subsection{Case Study in EB-Habitat}
\label{sec:case_study_habitat}

\begin{figure*}[!t]
    \centering
    \includegraphics[width=0.49\textwidth]{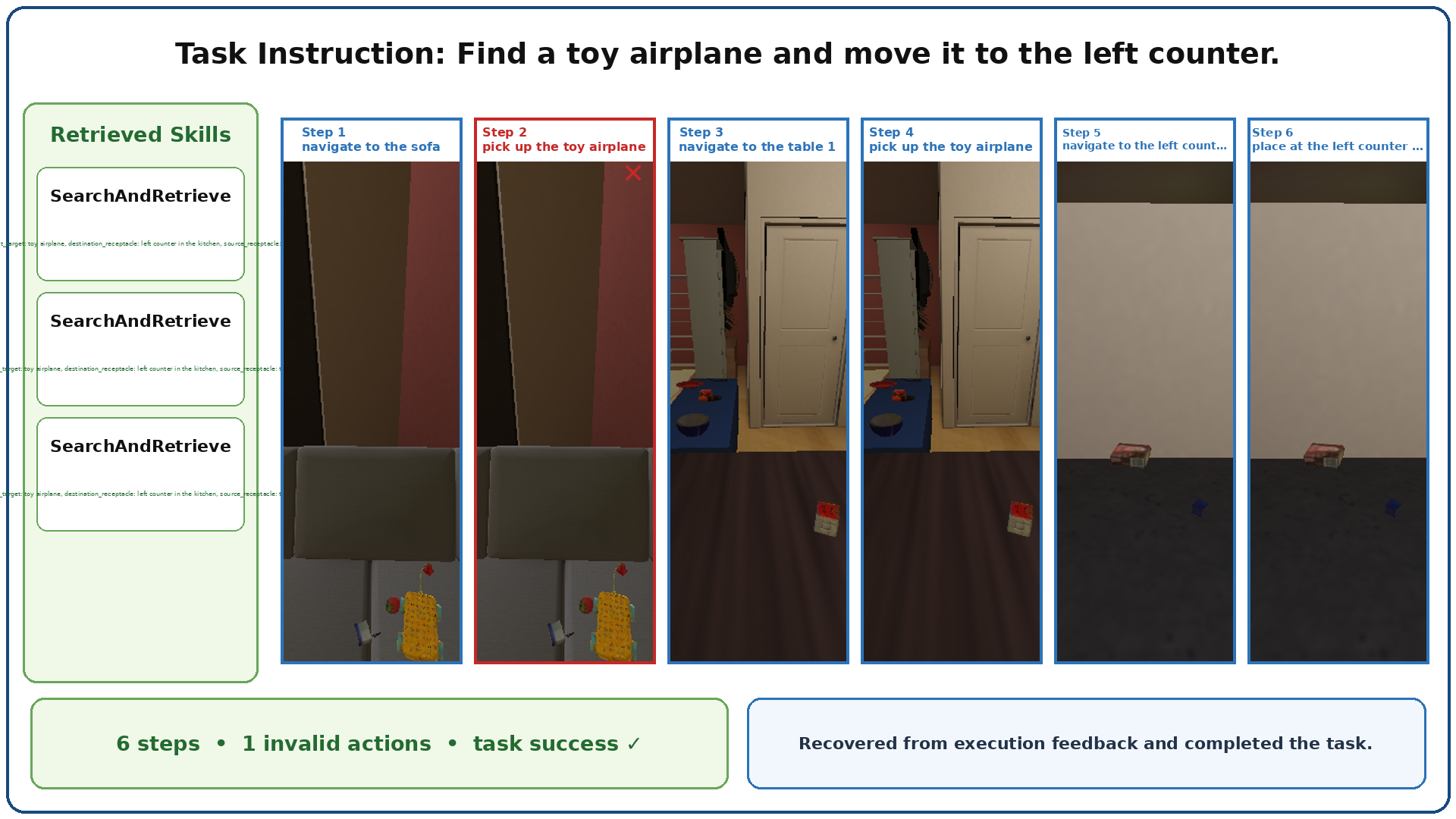}
    \hfill
    \includegraphics[width=0.49\textwidth]{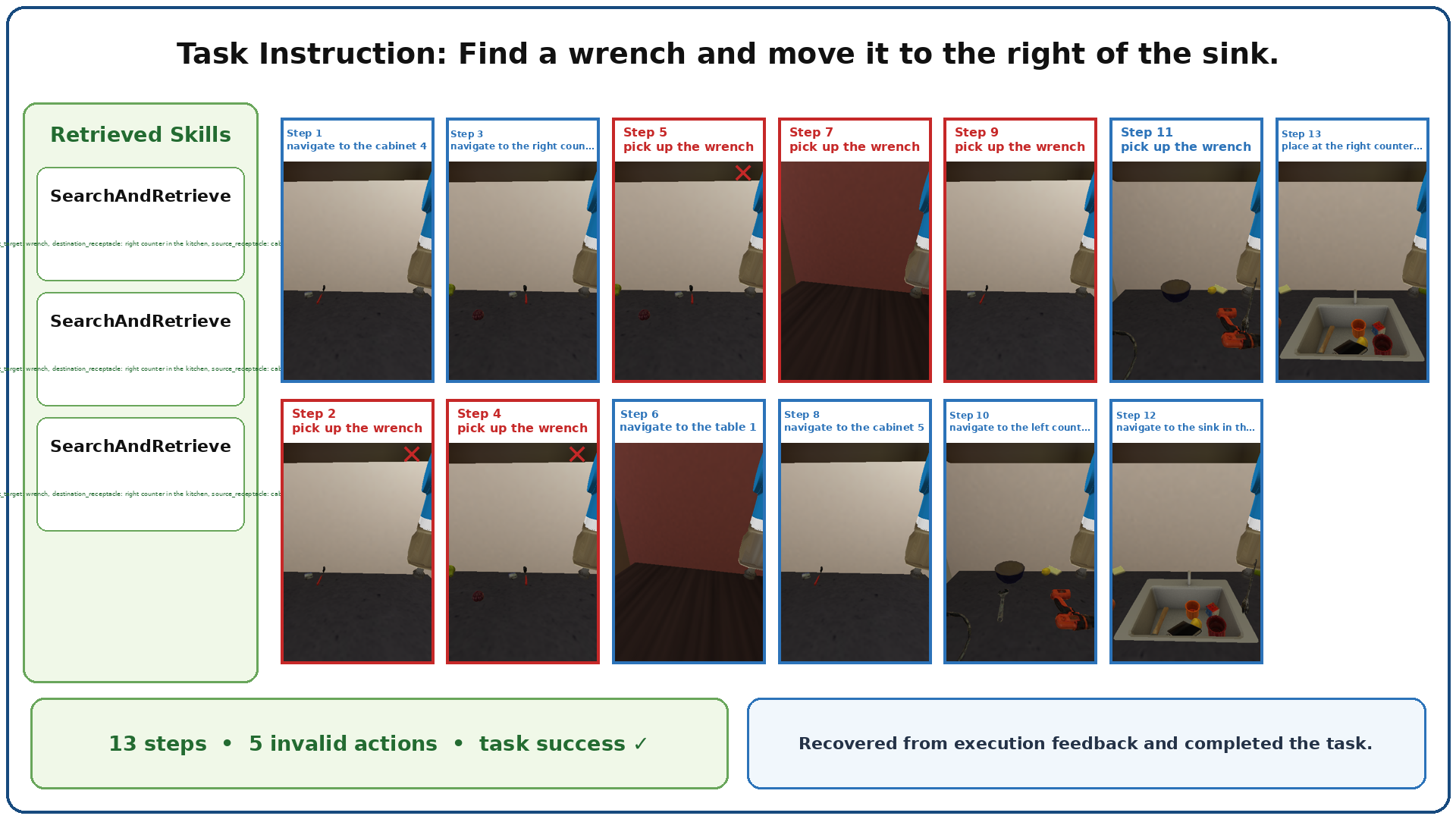}
    \caption{Feedback-guided object search in EB-Habitat. \textsc{SearchAndRetrieve} maintains the task structure while the executor switches candidate source receptacles.}
    \label{fig:habitat_search_recovery}
\end{figure*}

\paragraph{Successful cases: feedback-guided search.}
Unlike EB-ALFRED, EB-Habitat permits navigation only to receptacles. Object retrieval therefore requires the agent to infer a source receptacle before executing a pickup. Figure~\ref{fig:habitat_search_recovery} shows how \textsc{SearchAndRetrieve} structures this process. In Base-36, the agent first searches the sofa for a toy airplane. After pickup fails, it moves to another candidate receptacle, successfully retrieves the airplane from a table, and delivers it to the left counter. Spatial-35 exhibits a longer search for a wrench: the executor visits several cabinets and counters, but preserves the same object and destination arguments until pickup succeeds. The composite skill provides a stable navigation--pickup--navigation--placement skeleton, while feedback is used to revise the uncertain source location.

\begin{figure*}[!t]
    \centering
    \includegraphics[width=0.49\textwidth]{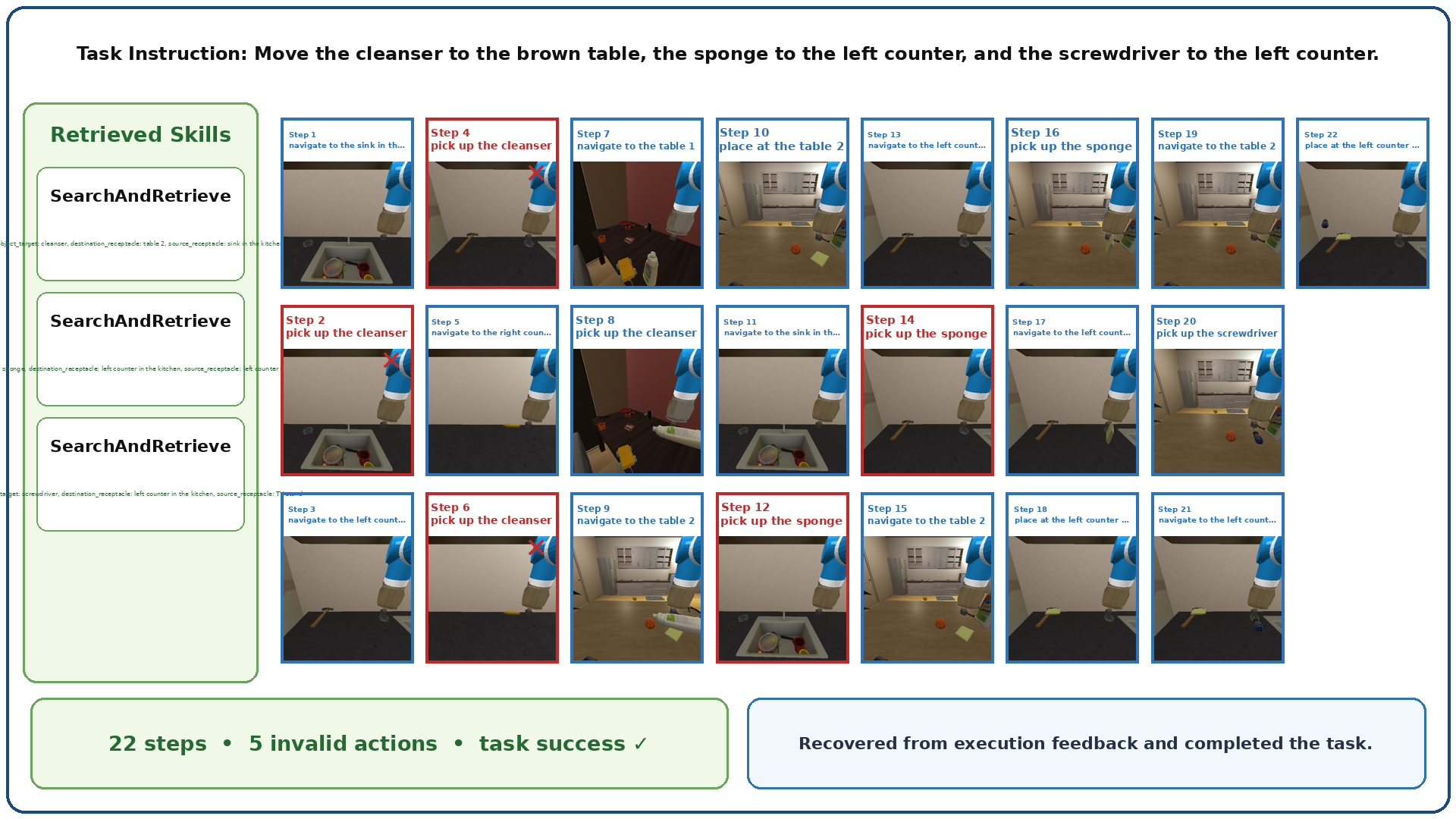}
    \hfill
    \includegraphics[width=0.49\textwidth]{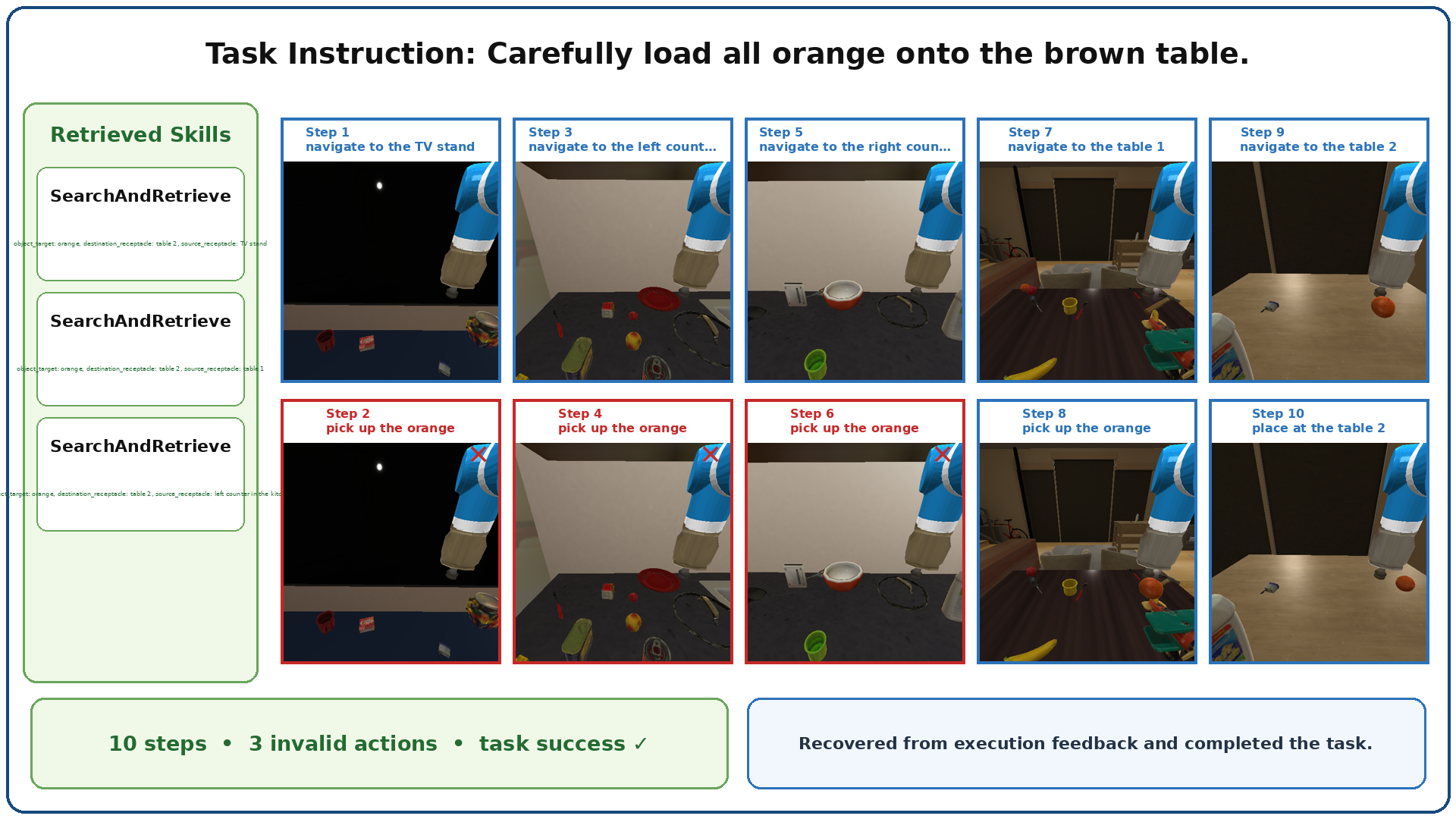}
    \caption{Successful multi-object transport in EB-Habitat. A shared \textsc{SearchAndRetrieve} structure is repeatedly instantiated for different objects and destinations.}
    \label{fig:habitat_multiobject}
\end{figure*}

\paragraph{Successful cases: repeated multi-object transport.}
The same abstraction also supports tasks containing several independent relocations. In Figure~\ref{fig:habitat_multiobject}, Long-14 sequentially moves a cleanser, sponge, and screwdriver to their respective destinations. Long-36 searches several receptacles and transports all orange instances to the brown table. Rather than generating a separate plan template for every object, the policy repeatedly instantiates \textsc{SearchAndRetrieve} with different object, source, and destination arguments. This reuse reduces planning complexity while still allowing each failed pickup to trigger local search.

\begin{figure*}[!t]
    \centering
    \includegraphics[width=0.49\textwidth]{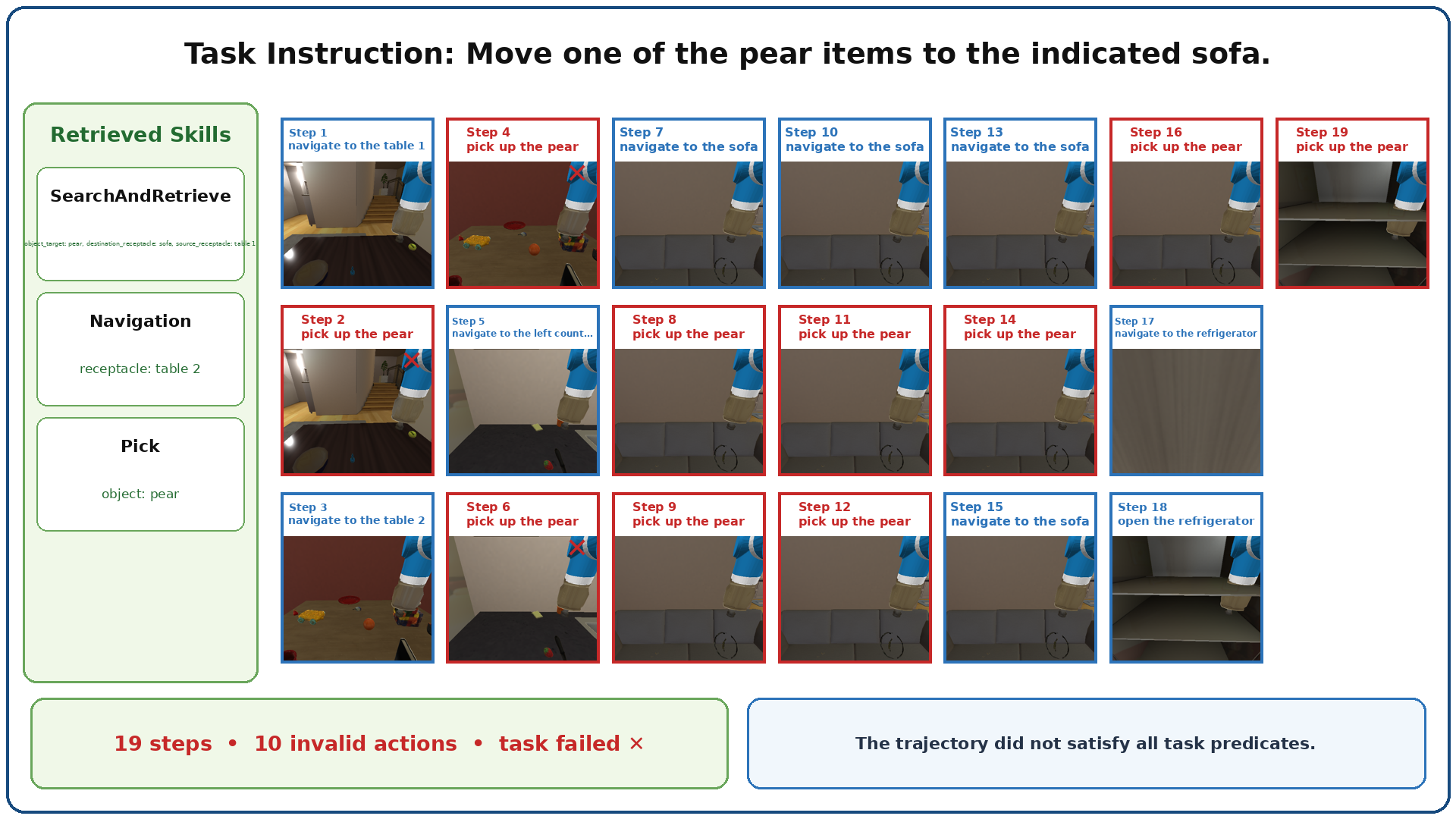}
    \hfill
    \includegraphics[width=0.49\textwidth]{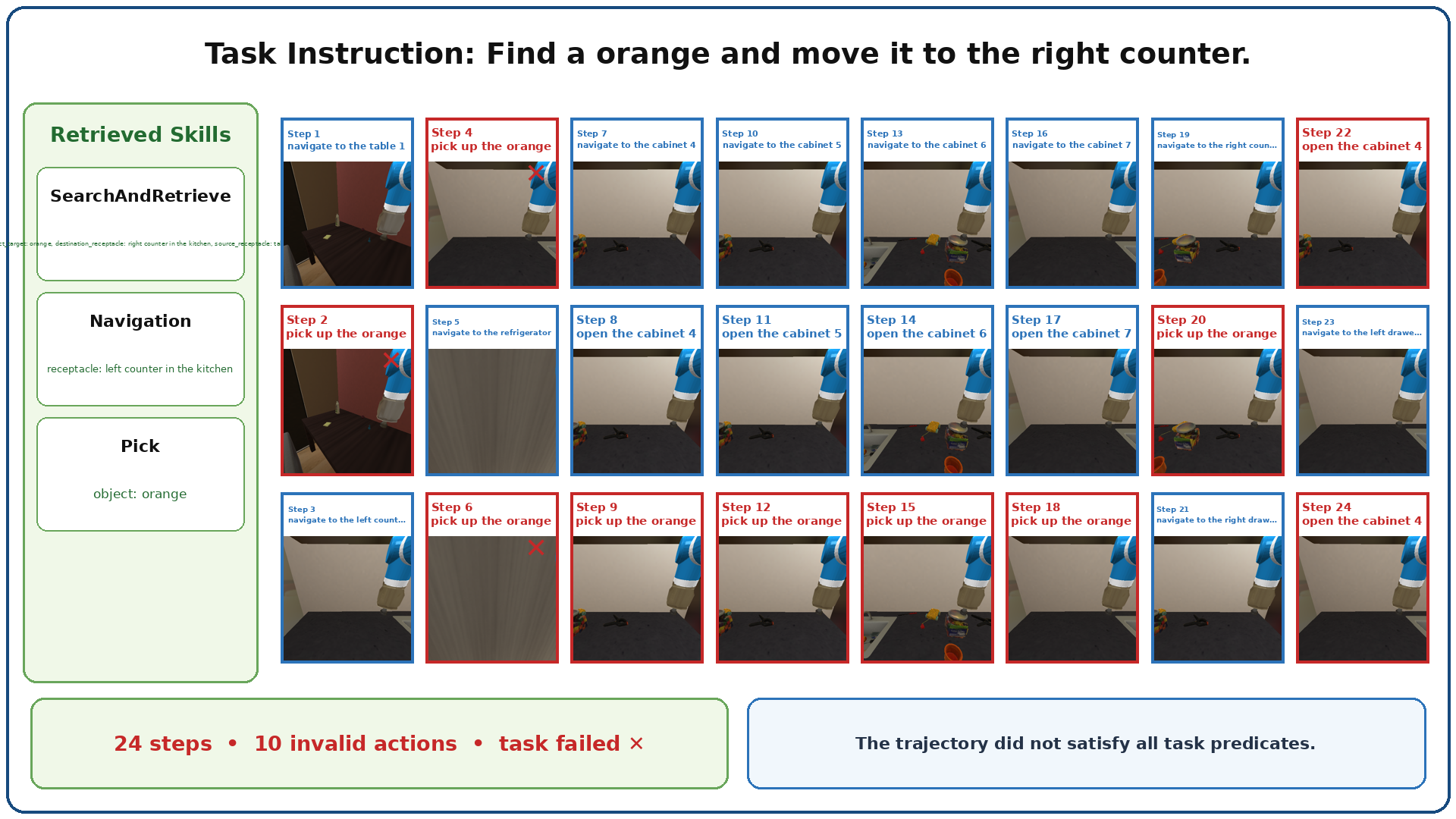}
    \caption{Source-receptacle failures in EB-Habitat. Incorrect initial hypotheses lead to repeated pickup attempts and unstructured exploration.}
    \label{fig:habitat_source_failures}
\end{figure*}

\paragraph{Failure cases: incorrect source-receptacle hypotheses.}
The same factor that makes \textsc{SearchAndRetrieve} useful also creates a Habitat-specific failure mode. In Figure~\ref{fig:habitat_source_failures}, Base-01 begins with an incorrect source hypothesis for the pear. Although the executor visits several receptacles, it repeatedly attempts pickup at locations where the object is not reachable and eventually explores unrelated containers. Base-26 similarly searches tables, counters, cabinets, and drawers for an orange, but lacks a coverage-aware strategy for deciding which receptacle to inspect next. The skill schema correctly represents object transport, but its source argument is uncertain; the eventual failure is primarily caused by executor-side exploration and recovery rather than by the navigation--pickup--placement pattern itself.

\paragraph{Shared and environment-specific observations.}
Both environments exhibit failures from semantic grounding, repeated invalid-action loops, multi-object identity tracking, and error accumulation over long trajectories. Their dominant bottlenecks, however, differ. EB-Habitat restricts navigation to receptacles, making object retrieval dependent on source-receptacle inference, indirect exploration, and the mapping of expressions such as ``right of the sink'' to simulator receptacles. Its long-horizon tasks mainly accumulate search costs across several relocations. EB-ALFRED permits more direct object search but places greater pressure on manipulation state, including holding constraints, object containment, and transformations such as slicing, washing, heating, and cooling.

Overall, skills in EB-Habitat are most valuable for structuring repeated object search and transport, whereas their main limitation is not the transport pattern itself but uncertain source-receptacle inference and ineffective exploration after failure. In contrast, EB-ALFRED is more strongly limited by nested manipulation, holding-state consistency, and long-horizon object-state transitions. Across both benchmarks, the learned library provides reusable planning structure, while robust grounding and state-aware execution remain essential for realizing its full benefit.

\end{document}